\documentclass[11pt]{article}
\usepackage[margin=1in]{geometry}
\usepackage[T1]{fontenc}
\usepackage{lmodern}
\usepackage{microtype}
\usepackage{amsmath,amssymb}
\usepackage{booktabs}
\usepackage{multirow}
\usepackage{graphicx}
\usepackage{xcolor}
\usepackage[round]{natbib}
\usepackage{url}
\usepackage{hyperref}
\hypersetup{colorlinks=true,linkcolor=blue,citecolor=blue,urlcolor=blue,
  pdftitle={Laya as a Typed Probabilistic Assessor: An Independent Reproduction and a Preregistered Study of Calibration and Selective Escalation},
  pdfauthor={Gowthamkumar Nandakishore}}

\newcommand{\meas}{\textsc{measured}}
\newcommand{\rep}{\textsc{reported}}
\newcommand{\tp}{\textsc{third-party}}

\newcommand{\descopedmark}{%
  \begin{center}%
  \fcolorbox{black}{gray!12}{\parbox{0.92\linewidth}{\centering\bfseries
  [DESCOPED: preregistration \S 0.a: no external-set results will be reported
  in this version. Conclusions are restricted to the synthetic benchmark.]}}%
  \end{center}}

\title{Laya as a Typed Probabilistic Assessor:\\
An Independent Reproduction and a Preregistered Study of\\
Calibration and Selective Escalation}
\author{Gowthamkumar Nandakishore}
\date{September 2026}

\begin{document}
\maketitle

\begin{abstract}
The shipped Laya Typed-Decisions checkpoint, a 421M-parameter ModernBERT-large
assessor that answers typed \texttt{choice}/\texttt{noul}/\texttt{score} questions
over workflow state, is uniformly \emph{under}-confident. The signed
confidence--accuracy gap is $-0.214$, every occupied reliability bin's accuracy
exceeds its confidence, and that sign uniformity collapses every binned ECE variant
to the same value, $0.214$. The card frames the risk as over-confidence; the
measured direction is the opposite, and the direction decides which way a
confidence-gated cascade fails. A single disjointly fitted temperature
($T{=}0.469$, sharpening) removes most of the miscalibration (held-out ECE $0.204
\to 0.037$) and outperforms the shipped per-option-count table. The frozen
selection rule instead chose isotonic regression, which overfit and failed its
held-out NLL contrast on both tracks, so hypothesis H2 is not supported. Re-running the released
checkpoint on its full official test split reproduces the card's headline accuracy
($0.767$ vs.\ $0.766$). The retrospective E1 reproduction preceded the analysis
freeze; E2--E8 were prospectively preregistered, and 20 of 22 executed confirmatory
tests reject under Benjamini--Hochberg FDR at $q{=}0.05$ (two descoped). The frozen
gate beats random escalation but misses its $10\%$ accepted-set error target on
both tracks, an exploratory out-of-distribution probe finds no zero-shot transfer
(accuracy $0.617$), and every score measures agreement with a synthetic teacher
whose self-agreement ceiling ($0.735$) the specialist exceeds. Per-decision
predictions, run manifests, and the frozen preregistration are in the
ancillary files.
The author has no affiliation with the model's publisher, the dataset's
publisher, or TypeSafe (\S\ref{sec:coi}).
\end{abstract}

\noindent\textbf{Keywords:} typed decisions; language-model evaluation; assessor;
calibration; selective prediction; reproducibility; preregistration

\section{Introduction}\label{sec:intro}

Automated assessment increasingly routes real decisions, such as whether an agent
trace needs human review, whether an invoice is paid or held, and whether a security
alert is closed or contained. A \emph{typed decision} interface serves this setting directly. The model
receives one piece of unstructured state and several typed questions, and returns a
probability distribution over each question's label space rather than free text. This
differs from reference-based answer grading, where a candidate is compared with a
supplied reference \citep{gisserot2026}, and from general verification, where a solution
or trajectory receives a graded quality score \citep{kwok2026}. It is closest in
interface to decision-only judging services \citep{li2026}\footnote{\citet{li2026}
evaluate the commercial TypeSafe JEV service as a black box; that study is
independent of TypeSafe and of this work.}, but the artifact studied
here is small, open-weight, and local.

The gap this paper addresses is the distance between a reported benchmark accuracy and a
deployable confidence signal. The public model card for Laya Typed-Decisions reports
$0.766$ accuracy on its benchmark's test split (\rep{}, model
card\footnote{\url{https://huggingface.co/convaiinnovations/laya-typed-decisions},
accessed 2026-09-26; revision \texttt{1a793eb5}.\label{fn:modelcard}}). Yet the same
card instructs users to treat the checkpoint's confidence as uncalibrated, and its
shipped temperature configuration is known to be misconfigured: the temperatures
fitted during training used a slice of the training data (\rep{}, model
card), and they are overridden at inference by an inherited per-option-count
table whose out-of-range \texttt{choice:11+} entry is clamped by \texttt{laya}
0.3.20 (\meas{}). Whether such an artifact can act as a trustworthy
first-stage assessor (accepting decisions when confident and escalating when unsure,
as \citet{li2026} frame it for a proprietary hosted judge) is an empirical question
that its card does not answer.

This is an independent study. The author has no affiliation with Convai
Innovations, which publishes the model, its base model, and the \texttt{laya}
package; with the LocalLLaMA organization, which publishes the benchmark; or
with TypeSafe (\S\ref{sec:coi}). What the study supplies is a frozen protocol,
frozen thresholds, and per-decision artifacts that let anyone check or extend
the numbers. A previously
circulated draft of this work was a protocol-only manuscript that proposed the
evaluation and reported no measurements. This version executes that protocol in two
registers. The reproduction (E1) ran first and is retrospective, predating the
preregistration under the plan's fit-nothing exemption. The plan was then
frozen, and every prospectively frozen analysis (E2--E8) executed against the
frozen thresholds; only those analyses are claimed as preregistered.
\S\ref{sec:prereg} states the timing, including the acceptance-gate item it fails,
and Appendix~\ref{sec:timeline} gives the full execution timeline. The one descoped
component is the external-validity set (E9; \S\ref{sec:e9}).

\paragraph{Contributions.}
\begin{enumerate}
  \item \textbf{A decomposition of the card's miscalibration (E4).} The shipped
  checkpoint is uniformly under-confident (signed gap $-0.214$), not over-confident as
  the card's guidance implies; one disjointly fitted temperature removes almost all of
  it (held-out ECE $0.204 \to 0.037$), and the fitted temperature \emph{sharpens}
  ($T{=}0.47$) where the shipped override softens ($T{=}1.25$--$1.98$). The
  inherited configuration is directionally backwards on this benchmark.
  \item \textbf{An escalation analysis with explicit scope limits (E5--E6).} No
  runnable strong judge exists in this study's environment (an author-directed
  constraint in the frozen plan), so cascade end-to-end accuracy is reported only as an
  \emph{oracle-escalation upper bound} with the escalation target's accuracy and cost
  as explicit symbolic parameters, alongside a random-escalation control at matched
  rate. The confidence signal does real work (E1t/E2t, $+6.5$/$+7.6$ accepted-set
  accuracy points over random escalation, both FDR-significant), yet the frozen gate
  misses its $10\%$ error target out-of-sample and passes confidently wrong security
  decisions (\S\ref{sec:e5}). The proprietary JEV service is never run (no API access
  exists) and appears only as third-party context with its source's own disclaimer.
  \item \textbf{A preregistered analysis protocol, frozen after the reproduction and
  before all fitting.}
  All thresholds, temperature-fitting procedures, split assignments, confidence
  statistics, hypothesis operationalizations, and a 24-test false-discovery-rate
  family were frozen (2026-09-26, 15:50~UTC) before
  any fitting touched the test split, and the frozen choices were then followed even
  where they hurt; the frozen method-selection rule picked an overfit calibrator on
  both tracks (\S\ref{sec:e4}). The design
  separates a split-half protocol on the released checkpoint (Track~A) from a clean
  re-training replication (Track~B), because the released checkpoint's calibration
  data is contaminated by construction (\S\ref{sec:splits}).
  \item \textbf{An independent reproduction of the released checkpoint (E1).} We
  re-run it on all 400 test cases and report every headline metric against
  the card's \rep{} values with 95\% BCa confidence intervals from a \emph{case-level}
  cluster bootstrap (the 2{,}000 decisions are five-per-case and correlated;
  decision-level intervals would be too narrow, Appendix~\ref{app:family}). Soft accuracy is the one headline
  metric that fails to reproduce ($0.509$ vs.\ $0.471$; CI excludes the reported
  value) and is recorded as an open reproducibility finding (\S\ref{sec:e1}). We
  release the per-decision prediction artifact (2{,}000 rows); the model card
  publishes no per-decision record behind its headline numbers. Per-decision
  artifacts for every headline run are released as well.
  \item \textbf{A clean replication that prices contamination (Track~B).} Re-training
  from the pinned base with the card's recipe, on case-level splits, on one
  Apple-silicon device, reproduces test accuracy within noise ($0.771$ vs.\ $0.767$;
  paired $p{=}0.57$). Its cleanly calibrated identity configuration already has
  ECE $0.151$ vs.\ the released $0.214$, while a probe that refits temperatures on
  trained-on items degrades held-out ECE $3.7\times$ ($0.024 \to 0.087$), a direct
  measurement of what the card's contaminated calibration protocol costs
  (\S\ref{sec:e7}).
\end{enumerate}

\section{Related work}\label{sec:related}

\paragraph{Reference-based encoder assessment.}
\citet{gisserot2026} show across 36 models and 15 tasks that lexical answer matching
correlates poorly with human judgment, and train an encoder (EuroBERT-210M;
\citealp{boizard2025}) on question--candidate--reference triplets to judge correctness
in reference-based settings. Two of their findings frame our study. Their
encoder's soft probabilities are usably calibrated after a single temperature
($\tau{=}1.75$). And their cost-effective \emph{fallback cascade}, regex
grading first with the encoder only where regex fails, cuts total compute roughly
$5\times$ when regex fails on 20\% of items (\tp{}; their benchmarks). Laya
Typed-Decisions descends from the same encoder lineage (it is a ModernBERT-large
checkpoint; \citealp{warner2024}), but its task is not reference-based; the benchmark
supplies no reference answers, so BERT-as-a-Judge is not runnable here and none of its
numbers can be compared to ours (preregistration \S0.2).

\paragraph{Confidence-gated cascades.}
Selective prediction descends from the reject option \citep{chow1970} through
selective deep networks \citep{geifman2017}. For language models, FrugalGPT sequences
cheaper models before expensive ones \citep{chen2024frugalgpt}, and
\citet{jitkrittum2023} characterize when confidence-based deferral is sufficient in
principle. Closest to us, JEV-as-a-Judge \citep{li2026} studies a
decision-only judge whose maximum label probability gates escalation to a stronger
generative judge. Their frozen cascade at threshold $0.9$ accepts $53.7\%$ of items and
retains ${\sim}99\%$ of the strong judge's accuracy at ${\sim}57$--$62\%$ of its fee
(\tp{}; their benchmarks). Three facts matter for reading this paper against
theirs. JEV is a \emph{proprietary hosted service} evaluated as a black box; its
benchmarks are five public judge suites, none of them the four workflow domains
studied here; and its authors have \textbf{never evaluated it on those domains}.
We run nothing against JEV (no API access exists; the frozen plan forbids any JEV
run or estimate) and make no head-to-head claim.

\paragraph{Fine-grained verification.}
LLM-as-a-Verifier \citep{kwok2026} computes continuous quality scores as expectations
over scoring-token logits and scales verification along granularity, repetition, and
criteria decomposition, reporting state-of-the-art results on agentic benchmarks
(Terminal-Bench~V2 $86.5\%$, SWE-Bench Verified $78.2\%$, RoboRewardBench $87.4\%$,
MedAgentBench $73.3\%$; \tp{}). Laya's interface is materially different (a trained
encoder emitting distributions over \emph{typed} label spaces rather than
scoring-token readout from a generative model), but the shared lesson is that
probability shape beyond the argmax carries the operational signal.

\paragraph{Calibration and selective prediction measurement.}
Modern networks are often miscalibrated and temperature scaling is the standard
post-hoc remedy \citep{guo2017}. Binned ECE is a biased estimator; bin placement
changes the number \citep{nixon2019,roelofs2022}, and recalibrated models can be less
calibrated than they report \citep{kumar2019}. We therefore report an ECE \emph{family} (equal-width at several
bin counts, equal-mass, classwise, and a signed confidence--accuracy gap) with proper
scoring rules \citep{gneiting2007} rather than a single ECE. Judge-style evaluators
also carry presentation biases, in particular position and order sensitivity
\citep{zheng2023judging,zheng2024mcq,pezeshkpour2024}, which motivates the
frozen robustness suite (E7). On
statistics, we follow bootstrap practice for clustered data
\citep{efron1993,davison1997}, Benjamini--Hochberg FDR control \citep{benjamini1995},
and preregistration norms \citep{nosek2018}.

\section{Setup}\label{sec:setup}

\subsection{Model}\label{sec:model}
Laya Typed-Decisions is a 421M-parameter typed-decision checkpoint built on
ModernBERT-large \citep{warner2024}, released by Convai Innovations
(footnote~\ref{fn:modelcard}). It answers three primitives: \texttt{choice}
(distribution over $N$ labels), \texttt{noul} (probability a statement is true), and
\texttt{score} (distribution over ordered rubric levels plus an expected score). Per
its card (\rep{}), it was fine-tuned from the \texttt{convaiinnovations/laya} base with
an RLCD-style recipe: the policy emits a distribution over typed options, and
exploration adds zero-mean Gaussian noise to logits ($\sigma$ annealed
$0.4\!\to\!0.1$). The reward
is a strictly proper scoring rule (log plus $0.75\cdot$spherical, plus a ranked
probability score term for ordinal \texttt{score} questions); updates are REINFORCE
with a group-mean baseline plus soft cross-entropy against teacher distributions. At
inference the shipped configuration applies a per-option-count temperature table
(\texttt{choice:3-5} $=1.760$, \texttt{noul:2} $=1.983$, \texttt{score:3-5} $=1.251$,
\ldots; \meas{}, read from the pinned checkpoint) which \emph{overrides} per-type
temperatures fitted during training, temperatures that were themselves fitted on a
slice of the training data (\rep{}, model card). The card's own
instruction is to treat shipped confidence as uncalibrated. We evaluate the checkpoint
exactly as shipped, bug and all, through \texttt{laya} 0.3.20 \texttt{predict\_batch}.

\subsection{Benchmark and the teacher-label caveat}\label{sec:benchmark}
The Typed Decisions benchmark\footnote{\url{https://huggingface.co/datasets/LocalLLaMA/typed-decisions},
accessed 2026-09-26; revision \texttt{f7a2487e}. The dataset is independent of, and
unaffiliated with, TypeSafe. The LocalLLaMA organization on Hugging Face that
publishes it is distinct from the Reddit community of the same
name.\label{fn:datasetcard}} has four workflow subsets
(\texttt{agent\_trace\_observability}, \texttt{customer\_service},
\texttt{invoice\_processing}, \texttt{security\_incidents}), each with 300 train and
100 test cases; every case asks five typed questions over one shared state, giving
6{,}000 train and 2{,}000 test decisions. Option \texttt{criteria} descriptions are
part of the input. Gold labels are the \emph{mean of three samples from a teacher
endpoint} (roughly 4B-class capability, temperature 0.7; \rep{}, dataset card); a
\texttt{label\_agreement} field records how much the samples disagreed.

Three dataset-card reference points (\rep{}) anchor interpretation: a majority
baseline scores $0.520$ (on the combined 1{,}600 cases), a model with perfect knowledge
of the latent generating factors scores $0.704$, and \emph{teacher self-agreement},
a fresh teacher sample scored against gold built from the others, scores $0.735$.
The card's own reading is that ``a score much above 0.75 means a model has learned the
teacher's quirks rather than the task.'' Agreement with a teacher is not correctness
\citep{gudibande2023}; the dataset card gives a concrete example where the teacher
missed a duplicate invoice. Every number in this paper measures teacher agreement on
synthetic cases, and \S\ref{sec:discussion} returns to what that does and does not
support. The card also separates \emph{specialist} models (fitted on these four
workflows) from \emph{generalist} ones (zero-shot on unseen schemas) and warns the two
are not comparable rows; we preserve that grouping everywhere.

\subsection{Split design (Track A and Track B)}\label{sec:splits}
The released checkpoint was fine-tuned on \emph{all} 1{,}200 training cases, so any
temperature or threshold fitted on training data is contaminated, the
defect its card admits. The frozen design therefore runs two
tracks and reports both.

\textbf{Track A (split-half test protocol; released checkpoint).} The 400 official
test cases are partitioned at case level, stratified by workflow (50/50 per workflow),
into \textsc{calib} (200 cases, 1{,}000 decisions) and \textsc{eval} (200 cases,
1{,}000 decisions), with generator and seed fixed in the preregistration (NumPy PCG64,
seed 20260926; ID lists in \texttt{track\_a\_split.json}, SHA-256
\texttt{7719b147\ldots}). Every temperature and threshold is fitted on \textsc{calib}
only; all Track~A headline results are reported on \textsc{eval} only. \textsc{eval}
results are not directly comparable to the card's 2{,}000-decision figures (half the
sample); the full-400 reproduction (E1) is reported separately and labeled as such,
not as a calibrated result.

\textbf{Track B (clean replication).} We fine-tune our own specialist from the pinned
\texttt{laya} base model with the card's RLCD recipe (\S\ref{sec:model}),
splitting the 1{,}200 training cases at case level into 960 fine-tune / 240 validation
(stratified, seed 20260926). Calibration and cascade thresholds are fitted only on the
240 validation cases; the 400-case test split is evaluated exactly once. Track~A
answers whether the released artifact can be used responsibly today; Track~B answers
whether the method works when the protocol is clean. Hypotheses H2/H3 are primarily
tested on Track~B.

\subsection{Comparators}\label{sec:comparators}
The preregistration (its \S0, an author-directed amendment recorded before the freeze)
restricts execution to models runnable on the local machine. The frozen \meas{}
comparator set:
\begin{itemize}\setlength{\itemsep}{1pt}
  \item Laya Typed-Decisions, released (Track~A);
  \item our Track~B replication;
  \item the \texttt{laya} base checkpoint zero-shot under the same question
  contract (card reports $0.362$; \rep{}, to verify);
  \item \texttt{laya-multilingual} (mmBERT-base, 322M; its card notes no fitted
  temperatures at all);
  \item a ModernBERT-base specialist trained by us with the Track~B recipe (the
  dataset card lists its own frozen-encoder specialist at $0.646$; \rep{});
  \item trivial baselines: majority class per question, uniform, training-set
  prior (model card reports majority $0.461$ and random $0.318$; dataset card
  reports uniform $0.308$ and prior $0.470$ on its own scoring; all \rep{});
  \item the teacher self-agreement ceiling, recomputed from the dataset's
  three-sample fields (card reports $0.735$; \rep{}).
\end{itemize}
The same directive dropped any API generative judge and any local ${\sim}$8B
generative judge.
Consequently \emph{no runnable escalation target exists}, with the consequences for
E5/E6 spelled out in \S\ref{sec:e5}. Comparators are grouped by task access and
training exposure in every table; a workflow specialist and a zero-shot generalist are
never ranked as interchangeable rows.

\textbf{JEV, third-party only.} The dataset card's maintainers measured TypeSafe JEV
1.13.0 through its public API on 2026-09-18 (all 400 test cases): accuracy $0.727$,
soft accuracy $0.580$, Brier $0.148$, ECE $0.144$ (signed over-confidence $+0.023$),
score MAE $0.391$, p50 latency 710\,ms/case, end-to-end through the public API, one
request at a time (\tp{}; footnote~\ref{fn:datasetcard}). This is the only JEV
latency figure verifiable at the pinned dataset-card revision; it is quoted with
that measurement's own scope and makes no intrinsic inference-time claim.
The Laya model card quotes JEV comparison figures with an explicit disclaimer, which
we adopt: \emph{these numbers were not measured in this work; prompts, pipeline, and
sample handling differ; no superiority claim in either direction is made from them}. JEV nearly saturates this
benchmark ($0.727$ against the teacher self-agreement ceiling, per the dataset
card), which further limits what any comparison could show.
The same card, at the pinned revision, also lists a hosted zero-shot model,
meraGPT Decider~1, at $0.768$ accuracy, Brier $0.052$, and score MAE $0.219$
(\tp{}; the card's own measurement dated 2026-09-22), above both JEV and the
teacher self-agreement ceiling. It is not run here (no API access) and appears
only as third-party context under the same disclaimer.

\subsection{Metrics}\label{sec:metrics}
All metric definitions are frozen in code (\texttt{metrics.py}; the metric code and
its unit tests are in the ancillary files). The 42 unit tests comprise 41 against
hand-computed cases plus one cross-check against the shipped
\texttt{laya.common.ece\_score}; 36 existed at the freeze and six were added with
the $\mathrm{ECE}_{\mathrm{sweep}}$ implementation under deviation D1. Metric
definitions are pinned to \texttt{laya/common.py} in the \texttt{laya} 0.3.20 PyPI
package and to the card's published evaluation
conventions. For decision $i$ with predicted distribution $p$ and
teacher distribution $g$ over $K$ options:
\begin{itemize}
  \item \textbf{Accuracy}: $\mathbf{1}[\arg\max_k p_k = \arg\max_k g_k]$, averaged
  over decisions; gold argmax follows the teacher-majority rule of the fine-tuning
  notebook linked from the model card (footnote~\ref{fn:modelcard}), ``the
  notebook'' hereafter.
  \item \textbf{Macro-F1}: per-class $F_1 = 2\,\mathrm{TP} / (2\,\mathrm{TP} +
  \mathrm{FP} + \mathrm{FN})$ over option positions, unweighted mean.
  \item \textbf{Soft accuracy}: $\sum_k p_k g_k$ after the notebook's $10^{-6}$
  missing-key floor and renormalization, averaged over \texttt{choice} and
  \texttt{noul} decisions \emph{only} ($n{=}1{,}200$ of 2{,}000); the notebook's
  \texttt{score} arm never contributes. This scope is extracted from the card's own
  code; the card's $0.471$ is this quantity.
  \item \textbf{Brier}: $\sum_k (p_k - y_k)^2$ averaged over decisions. The card's
  $0.062$ takes $y$ to be the \emph{teacher's soft distribution} and averages over
  \texttt{choice}+\texttt{noul} only (verified against the notebook's code); we
  compute both this form and the one-hot-gold form. The frozen plan
  misattributed the card's figure to the one-hot form; deviation D6
  (Appendix~\ref{app:deviations}) records the correction, which changes no measured
  value.
  \item \textbf{NLL}: $-\operatorname{mean}\log\max(p_{\text{gold}}, 10^{-12})$.
  \item \textbf{RPS} (ordinal \texttt{score} questions):
  $\sum_k (\mathrm{CDF}_p(k) - \mathrm{CDF}_y(k))^2 / (K-1)$, matching the training
  reward's form.
  \item \textbf{ECE family}: the card's convention as frozen (10 equal-width bins;
  first bin closed at the bottom, others $(\mathrm{lo},\mathrm{hi}]$; empty bins
  skipped; mass-weighted) for card comparability. Also equal-width with
  $M \in \{10,15,20\}$, of which $M{=}15$ is the frozen confirmatory endpoint
  (a draft mislabeling is reconciled as deviation D7, Appendix~\ref{app:deviations},
  with no numeric effect);
  equal-mass (adaptive) binning \citep{nixon2019}; and classwise calibration error
  over option positions. The \textbf{signed gap} $=$ mean confidence $-$ accuracy
  shows the \emph{direction} ECE hides. $\mathrm{ECE}_{\mathrm{sweep}}$
  \citep{roelofs2022} is in the frozen family; its unit-tested implementation landed
  slightly after the E4 \textsc{calib} extraction began (deviation D1,
  Appendix~\ref{app:deviations}) and it is reported wherever the ECE family appears.
  \item \textbf{Error-detection AUROC/AUPRC}: errors as the positive class, score
  $-q$ where $q = \max_k p_k$; plus \textbf{AURC} for selective prediction.
\end{itemize}
The frozen confidence statistic is $q = \max_k p_k$ of the post-calibration
distribution; entropy, margin, and ordinal variance are secondary ablations only.

\subsection{Uncertainty quantification and multiplicity}\label{sec:stats}
The 2{,}000 test decisions sit in 400 case clusters (five decisions per case, same
document and context), so decisions are not independent. Every confidence interval in
this paper is a \textbf{case-level cluster bootstrap}: resample cases with
replacement, keep each case's five decisions together, $n_{\mathrm{boot}} = 10{,}000$,
seed 0; BCa intervals primary for single-quantity metrics, percentile as robustness
check \citep{efron1993,davison1997}. Model comparisons use paired differences on
identical resampled case sets; paired-difference intervals are percentile (BCa on a
difference needs a combined jackknife and was not preregistered for the paired case).
Bootstrap p-values use the finite-sample add-one convention $(b{+}1)/(B{+}1)$, doubled
for two-sided tests, so zero exceedances in $B{=}10{,}000$ resamples is reported as
$p \le 10^{-4}$ (one-sided) or $p \le 2{\times}10^{-4}$ (two-sided), the resolution
limit instead of zero. Naive i.i.d.\ decision-level intervals appear only in a
frozen appendix contrast (Figure~\ref{fig:f8}). Confirmatory inference is
limited to a frozen family of 24 tests with Benjamini--Hochberg FDR control at
$q = 0.05$ \citep{benjamini1995}; everything else is exploratory and labeled as such.
Under the preregistration's \S0.a scope amendment, the two external-validity family
members (F1t/F2t) could not execute in this version and are reported as not executed;
the BH denominator conservatively remains the full family size of 24 (equivalent to
assigning them $p{=}1$). Twenty of the 22 executed tests reject
(Appendix~\ref{app:family}).

\subsection{Preregistration and test-set exposure}\label{sec:prereg}
The analysis plan was frozen on 2026-09-26 at 15:50~UTC, before any fitting touched
the test split \citep{nosek2018}; the frozen document is released verbatim as an
ancillary file and its SHA-256 is listed in \texttt{anc/README.txt}. Timestamps are
taken from the run manifests, which are also released. The freeze time itself is
the author's own statement; the frozen document dates its own authorization to
21:16~IST (15:46~UTC) on 2026-09-26, and the first post-freeze manifest is
timestamped 15:51:06~UTC,
which brackets it. E1
predates the freeze. The preregistration (its \S2) quotes the plan clause that
exempts the reproduction because it fits nothing (no temperature,
threshold, prompt, or model choice was informed by test labels). But the same plan's
higher-priority integrity rule, that the official test split is opened only after
all choices are frozen, and its acceptance-gate wording (preregistration frozen
before the first test evaluation) admit no exemption. The two plan clauses conflict; the
conflict was resolved in favor of running E1 first, so E1 and H1 are retrospective,
only E2--E8 are prospectively preregistered, and the acceptance gate's timing item is
recorded as a disclosed failure in \texttt{gate.json}.
Post-freeze test-set exposure, exhaustively:
\begin{itemize}\setlength{\itemsep}{1pt}
  \item one accidental partial smoke re-run of the exempt E1 pipeline (logged;
  Appendix~\ref{app:deviations});
  \item the \textsc{calib} fitting pass (E4);
  \item two failed E4 attempts at 16:05:29 and 16:05:59~UTC (Python exceptions;
  manifested; read \textsc{calib} cases before failing);
  \item the single frozen \textsc{eval} pass (E5--E6);
  \item Track~B's single 400-case test pass;
  \item one full-400 pass per frozen comparator (each exactly once, manifested);
  \item E7's perturbation forwards, which run on \textsc{calib} cases only;
  \item the 2026-09-27 fp32/CPU rerun of all 400 test cases (fits nothing;
  deviation D10, Appendix~\ref{app:deviations});
  \item the 2026-09-27 version-bisection reruns of all 400 test cases, one per
  evaluated \texttt{laya} release (0.3.3 and 0.3.20), plus one single-case API
  schema check under 0.3.3 (all fit nothing; deviation D11).
\end{itemize}
E8 consumes only
records stored by the single \textsc{eval} pass; no new \textsc{eval} forwards occurred
after step 8. Hypothesis H1's numerical criterion ($|\Delta| \le 0.02$) was articulated
after E1 ran and is therefore a post-hoc reporting structure for E1 rather than a
preregistered prediction; H2--H4 and all thresholds were frozen before any
\textsc{calib}/\textsc{eval} or Track~B access. Every raw-logit extraction used by
E4/E5 was cross-checked against the E1 \texttt{predict\_batch} predictions
(2{,}000/2{,}000 decisions, max probability difference $1.2\times10^{-4}$, within the
stored files' 4-decimal rounding).

\section{Reproduction of the released checkpoint (E1)}\label{sec:e1}

E1 runs the released checkpoint, as shipped (including the inherited
\texttt{temperature\_by\_options} override), on all 400 test cases / 2{,}000 decisions:
\texttt{laya} 0.3.20 \texttt{predict\_batch}, one case per call, max sequence length
1024, batch size 1, Apple M5~Max (64\,GB, \texttt{mps}), Python 3.12.14, torch 2.14.0,
transformers 5.17.0. Every number labeled \meas{} below comes from the logged,
manifested run whose artifacts are released in the ancillary files; \rep{} numbers
are transcribed from the model card (footnote~\ref{fn:modelcard}). Nothing was tuned
toward the reported values.

\begin{table}[t]
\centering\footnotesize
\setlength{\tabcolsep}{4.5pt}
\begin{tabular}{lrrrl}
\toprule
Metric & Card (\rep{}) & Ours (\meas{}) & $\Delta$ & 95\% CI (BCa, \meas{}) \\
\midrule
Accuracy (all, $n{=}2000$)        & 0.766 & 0.7670 & $+0.0010$ & $[0.7455, 0.7870]$ \\
Soft accuracy (ch+no, $n{=}1200$) & 0.471 & 0.5086 & $+0.0376$ & $[0.4987, 0.5196]$ \\
Brier vs.\ teacher (ch+no)        & 0.062 & 0.0658 & $+0.0038$ & $[0.0610, 0.0716]$ \\
ECE (card conv., 10 bins)         & 0.213 & 0.2142 & $+0.0012$ & $[0.1954, 0.2328]$ \\
Score MAE (\texttt{score}, $n{=}800$) & 0.242 & 0.2424 & $+0.0004$ & $[0.2272, 0.2580]$ \\
\bottomrule
\end{tabular}
\caption{\textbf{E1 headline reproduction (T1).} \rep{} = transcribed from the model
card (accessed 2026-09-26); \meas{} = this work's logged run (manifest
\texttt{20260926T120855Z}). $\Delta$ = measured $-$
reported. Intervals: case-level cluster bootstrap (400 clusters, $n_{\mathrm{boot}} =
10{,}000$, BCa). ``ch+no'' = \texttt{choice}+\texttt{noul} scope, which is the card's
own scope for those metrics (\S\ref{sec:metrics}).}
\label{tab:t1}
\end{table}

\begin{table}[t]
\centering\footnotesize
\setlength{\tabcolsep}{4.5pt}
\begin{tabular}{lrrrl r}
\toprule
Slice & Card (\rep{}) & \meas{} & $\Delta$ & 95\% CI (BCa, \meas{}) & $n$ \\
\midrule
\texttt{agent\_trace\_observability} & 0.730 & 0.7300 & $0.0000$ & $[0.6800,\ 0.7740]$ & 500 \\
\texttt{customer\_service}           & 0.764 & 0.7660 & $+0.0020$ & $[0.7206,\ 0.8058]$ & 500 \\
\texttt{invoice\_processing}         & 0.804 & 0.8060 & $+0.0020$ & $[0.7660,\ 0.8380]$ & 500 \\
\texttt{security\_incidents}         & 0.766 & 0.7660 & $0.0000$ & $[0.7260,\ 0.8000]$ & 500 \\
\midrule
\texttt{noul} accuracy   & 0.857 & 0.8583 & $+0.0013$ & n/a & 600 \\
\texttt{choice} accuracy & 0.733 & 0.7333 & $+0.0003$ & n/a & 600 \\
\texttt{score} accuracy  & 0.723 & 0.7238 & $+0.0008$ & n/a & 800 \\
\texttt{noul} ECE   & 0.192 & 0.1938 & $+0.0018$ & n/a & 600 \\
\texttt{choice} ECE & 0.255 & 0.2548 & $-0.0002$ & n/a & 600 \\
\texttt{score} ECE  & 0.199 & 0.1991 & $+0.0001$ & n/a & 800 \\
\bottomrule
\end{tabular}
\caption{\textbf{E1 per-workflow and per-primitive reproduction.} Same provenance and
interval conventions as Table~\ref{tab:t1}. Note that \texttt{choice} ECE ($0.255$) is
\emph{worse} than the aggregate ECE ($0.214$); the aggregate flatters the model's most
error-prone primitive.}
\label{tab:t2}
\end{table}

\subsection{Headline metrics match the card, except soft accuracy}
Accuracy, Brier, ECE, score MAE, all four per-workflow accuracies, and all six
per-primitive statistics reproduce the card within $+0.004$
(Tables~\ref{tab:t1}--\ref{tab:t2}); the argmax pipeline therefore matches the card's, and H1 is supported on those metrics under the (post-hoc,
\S\ref{sec:prereg}) $|\Delta| \le 0.02$ criterion. Candidate failure causes
named in the frozen plan for investigation (tokenizer version, batch-size padding effects,
option-count clamping) were not pursued further for these metrics because the
deltas were trivial.

\paragraph{Open reproducibility finding 1, narrowed twice: soft accuracy does not
reproduce; precision, backend, and the released harness range are ruled out.}
Measured $0.5086$ $[0.4987, 0.5196]$ against a reported $0.471$. The interval
excludes the reported value, so this is a pipeline difference rather than sampling noise.
Because accuracy, ECE, and Brier all reproduce closely, the predicted distributions
must be close to the card's; a distribution slightly more concentrated on the
teacher's mode raises $\sum_k p_k g_k$ while barely moving $\sum_k (p_k - g_k)^2$ or
max-probability calibration. Two exploratory reruns of the same 2{,}000 decisions
narrow the cause. An fp32/CPU rerun (deviation D10; artifact
\texttt{e1\_fp32\_cpu\_probe.json}) reproduces the canonical run (F16 weights
under bf16 autocast on MPS) almost
exactly (soft accuracy $0.5087$; maximum per-decision probability difference
$0.0079$; 4 argmax flips of 2{,}000), ruling out numeric precision and the MPS
backend. A version bisection of the \texttt{laya} inference harness then reran the
decisions in one fresh virtual environment per release (deviation D11; artifacts
\texttt{e1\_version\_probe\_*.json}). Torch and transformers were pinned to the
canonical versions in every environment. \texttt{laya} 0.3.20 reproduces the
canonical run byte-identically at the stored 4-decimal rounding (maximum
probability difference $0.0$). \texttt{laya} 0.3.3, the lower endpoint of the
planned bisection range
(released 2026-09-19, single-state \texttt{predict} API), measures soft
accuracy $0.5087$. The two endpoints agree, so under the pre-stated rule no
intermediate release was bisected: every released harness version from 0.3.3 to
0.3.20 is ruled out. What remains is the card-side pipeline itself (its exact
package pins are not recoverable from the card): either its unreleased or
pre-0.3.3-era evaluation code or a different data state produced $0.471$. The
finding stays open, narrowed to the card side, with the per-decision files
released so the card's maintainers can close it.

\subsection{The shipped checkpoint is under-confident, not over-confident}
\label{sec:underconf}

\begin{figure}[t]
\centering
\includegraphics[width=\linewidth]{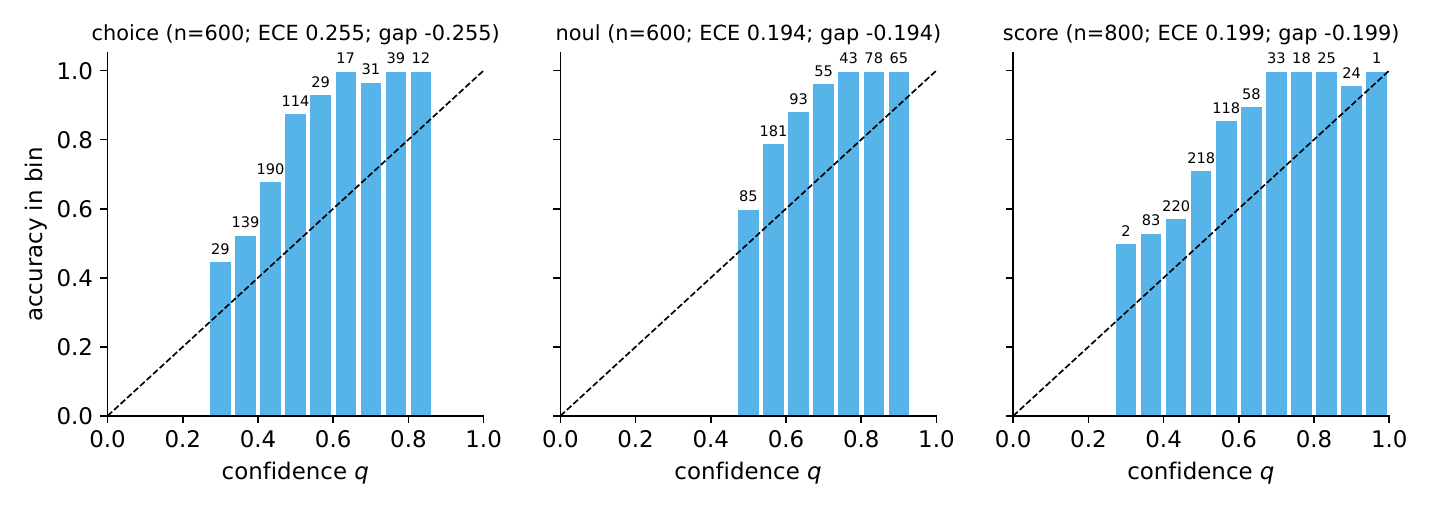}
\caption{\textbf{F1} (\meas{}). \textbf{Per-primitive reliability diagrams for the released
checkpoint on the full official test split} (E1; 2{,}000 decisions; Apple M5~Max,
MPS). Confidence is $q = \max_k p_k$ of the shipped
(post-temperature) distribution; binning is 15 equal-width bins (the frozen
confirmatory convention; on this run every binned ECE variant, including the card's
10-bin convention, collapses to the same value; only non-empty bins drawn; per-bin
decision counts annotated). Every occupied
bin in every primitive lies \emph{above} the diagonal; the shipped checkpoint is
uniformly under-confident. Recomputed per-primitive ECEs from the released
per-decision artifact match the logged E1 metrics exactly.}
\label{fig:f1}
\end{figure}

The signed calibration gap is $-0.2142$ (\meas{}), mean confidence $0.553$ against
accuracy $0.767$. Figure~\ref{fig:f1} shows that \emph{every} occupied reliability bin,
in every primitive, sits above the diagonal (per-bin gaps $-0.065$ to $-0.275$ over
1{,}767 distinct confidence values). This explains an otherwise puzzling measurement.
The entire binned-ECE family collapses to a single value, $0.21423$, for equal-width
binning at $M \in \{10,15,20\}$ \emph{and} equal-mass binning (\meas{}); when every
bin's gap has the same sign, mass-weighted binned ECE reduces to
$|\text{mean confidence} - \text{accuracy}|$ regardless of binning scheme, a known
failure mode of reading a single binned ECE \citep{nixon2019,roelofs2022,kumar2019}.

The \emph{direction} also contradicts the card's framing. The card instructs users
to treat the checkpoint's confidence as uncalibrated (guidance written around its
known temperature-configuration defect), and its comparison table reads as an
over-confidence concern; the dataset card separately reports JEV's signed gap as
over-confident ($+0.023$; \tp{}). The shipped Laya checkpoint measured here errs in
the \emph{opposite} direction. The inherited override applies only softening
temperatures on this benchmark (all used buckets $>1$: $1.760$, $1.983$, $1.251$;
\meas{}), which divide the logits and depress $q$ uniformly. The magnitude of the
card's ECE ($0.213$) is real, but its sign, on the shipped artifact, is
under-confidence. Whether the \emph{raw pre-temperature} logits are over-confident,
i.e.\ how much of it is configuration artifact versus model, is
the frozen E4 question (\S\ref{sec:e4}).
For cascade deployment the direction
matters. Uniform under-confidence inflates escalation rates (costly) rather than
concealing errors (unsafe); the measured operating consequences appear in \S\ref{sec:e5}.

\subsection{The temperature clamp is a no-op on this benchmark}
The shipped override contains one out-of-range entry (\texttt{choice:11+} $= 0.1006$,
which \texttt{laya} 0.3.20 clamps to $0.5$; \meas{}). No test-split \texttt{choice} question
has more than 5 options, so that bucket is never selected, and every bucket that is
used is inside the clamp range (\meas{}; clamp changes no used bucket). Running
through \texttt{predict\_batch} therefore reproduces the raw shipped configuration
exactly here; the well-known \texttt{choice:11+} sharpening bug does \emph{not} drive
the card's headline ECE, and would only bite deployments with $>$10-option questions.

\subsection{Supplementary measured statistics and latency}
On the full split (\meas{}, exploratory): NLL $0.7069$ ($\epsilon = 10^{-12}$);
macro-F1 $0.7499$; RPS on \texttt{score} questions $0.0795$ against one-hot gold and
$0.0156$ against the teacher distribution; error-detection AUROC $0.7584$ and AUPRC
$0.4423$ (errors positive, score $-q$). For context only, \citet{li2026} report JEV
error-detection AUROC of $0.869/0.745/0.863$ on \emph{their} three benchmarks and
near-chance $0.518$ on reference-free prose (\tp{}; different tasks, so no comparison is drawn).
Whether $0.758$ is enough ranking signal for a useful cascade here is measured in
\S\ref{sec:e5}.

Latency (\meas{}, this hardware; the card's T4 figures are on different hardware): batch size
1, first 3 cases discarded as warm-up; per-case p50/p90/p99 $= 95.9 / 124.1 / 195.4$\,ms
(five decisions per case; p50 per decision $19.2$\,ms; $54.3$ decisions/s). A batched
throughput sweep measured $51.9 / 42.9 / 37.8 / 38.7$ decisions/s at batch sizes
$1/8/32/64$; batching does \emph{not} improve throughput on this MPS path (an
unexplained, hardware-specific observation reported as-is; the card's T4 claim of
103--332 questions/s batched is \rep{} and untested here).

\subsection{Released artifact}
E1's per-decision predictions are released as
\texttt{released\_full\_test.jsonl} (2{,}000 rows; schema in
Appendix~\ref{app:artifact}); the model card publishes no per-decision record
behind its headline numbers, so this artifact makes the card's headline
independently re-analyzable, including every analysis in this paper, without
model access. Canonical copies of all E1 artifacts are in the ancillary files.

\section{Accuracy and probability quality across comparators (E2--E3)}\label{sec:e2}

\begin{figure}[t]
\centering
\includegraphics[width=0.75\linewidth]{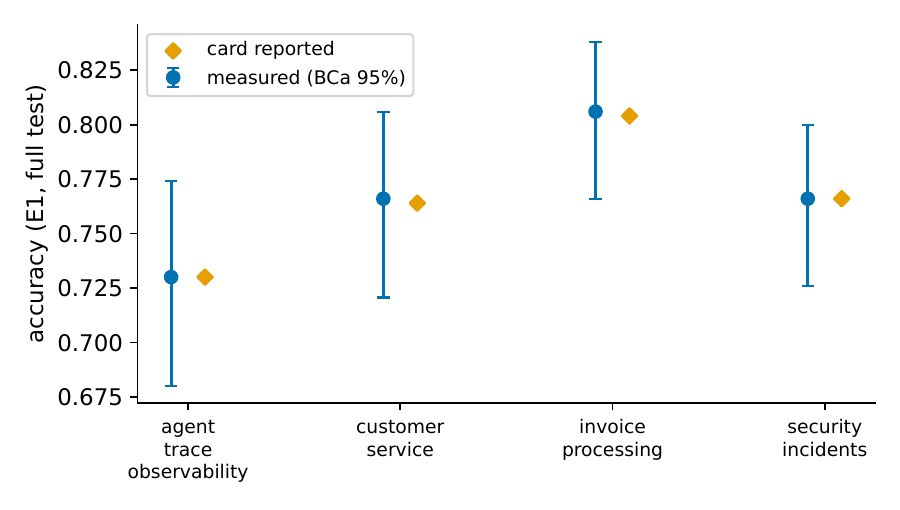}
\caption{\textbf{F5} (\meas{}). Per-workflow E1 accuracy, measured (BCa 95\%
case-cluster intervals) vs.\ card-reported points. All four workflows reproduce within
$0.002$.}
\label{fig:f5}
\end{figure}

All comparators of \S\ref{sec:comparators} were scored on the full 400-case test split,
each exactly once, each with a run manifest; Table~\ref{tab:t3} groups them by task
access and training exposure, per the dataset card's own discipline. Trivial baselines
are fitted on the 1{,}200 \emph{training} cases only. Figure~\ref{fig:f5} shows the
per-workflow reproduction with clustered intervals.

\begin{table}[t]
\centering\scriptsize
\setlength{\tabcolsep}{3pt}
\begin{tabular}{llrrrl}
\toprule
Group & Model & Acc.\ (\meas{}) & NLL & ECE$_{15}$ & 95\% BCa CI (acc.) \\
\midrule
\multirow{3}{*}{specialist}
 & Laya Typed-Decisions (released)        & 0.7670 & 0.7069 & 0.2142 & $[0.7455, 0.7870]$ \\
 & Laya Track~B replication (ours)        & 0.7710 & 0.6271 & 0.1509 & $[0.7495, 0.7905]$ \\
 & ModernBERT-base specialist (ours)      & 0.5425 & 0.9736 & 0.0728 & $[0.5145, 0.5685]$ \\
\midrule
\multirow{2}{*}{generalist, zero-shot}
 & \texttt{laya} base                     & 0.3630 & 1.3311 & 0.1731 & $[0.3385, 0.3875]$ \\
 & \texttt{laya-multilingual}             & 0.3495 & 1.8616 & 0.3159 & $[0.3270, 0.3717]$ \\
\midrule
\multirow{3}{*}{trivial (fit on train)}
 & majority per question                  & 0.4835 & 14.27  & 0.5165 & $[0.4585, 0.5070]$ \\
 & training-set prior                     & 0.4785 & 1.0344 & 0.0336 & $[0.4505, 0.5045]$ \\
 & uniform (analytic expectation)         & 0.3175 & 1.2118 & 0.0000 & $[0.3147, 0.3202]$ \\
\midrule
\multirow{3}{*}{reference}
 & teacher self-agreement (recomputed)    & \multicolumn{4}{l}{$0.594$ floor -- $0.865$ estimate; card: $0.735$ (\rep{})} \\
 & TypeSafe JEV 1.13.0                    & \multicolumn{4}{l}{$0.727$ (\tp{}; different pipeline; no comparison claimed)} \\
 & meraGPT Decider~1                      & \multicolumn{4}{l}{$0.768$ (\tp{}; card's measurement; no comparison)} \\
\bottomrule
\end{tabular}
\caption{\textbf{Comparator results on the full official test split (T3).} All
\meas{} rows are this work's logged runs, 2{,}000 decisions each, accuracy CIs from
case-level cluster bootstrap (BCa, $n_{\mathrm{boot}}{=}10{,}000$). NLL uses the
one-hot gold argmax; the majority baseline's NLL is degenerate by construction
(near-one-hot predictions). Specialists and zero-shot generalists are \emph{not}
comparable rows (dataset-card discipline). The teacher ceiling is recomputed from the
shipped \texttt{label\_agreement} fields, which record agreement but not the three raw
samples; the strict all-agree rate ($0.594$, \meas{}) is a floor, and assuming every
disagreement is a 2--1 split gives $0.865$ (\meas{}); the card's $0.735$ (\rep{}) lies
inside this bracket and could not be recomputed exactly without the raw samples.}
\label{tab:t3}
\end{table}

The zero-shot generalists confirm their cards (all \meas{}).
The \texttt{laya} base measures $0.3630$ against a reported $0.362$, and
\texttt{laya-multilingual} ($0.3495$), which ships no fitted temperatures at all, has
the worst calibration in the table (ECE $0.316$); it sits below the majority baseline
on accuracy, consistent with these schemas being unseen. The trivial baselines
measure within $0.023$ of their card values (majority $0.4835$ vs.\ reported $0.461$
and prior $0.4785$ vs.\ $0.470$, both slightly above; uniform $0.3175$ vs.\ random
$0.318$, slightly below). The first two
plausibly differ by fitting split and tie-breaking convention, and are reported as our
own measured definitions (majority/prior per (workflow, question) over training gold).
The specialist--teacher gap is the dominant structure. Both large-encoder
specialists sit at the teacher's own self-agreement region, at least $0.28$ above the
strongest non-specialist row (majority, $0.4835$) and up to $0.45$ above the weakest
(uniform, $0.3175$). The interesting differences among them are in
probability quality (NLL, ECE) rather than accuracy, which is where the released
checkpoint's inherited configuration hurts it (Track~B's clean identity beats it by $0.080$ NLL and $0.063$
ECE at indistinguishable accuracy; \S\ref{sec:e7}).

\paragraph{The RLCD recipe on a ModernBERT-base encoder does not reach the dataset
card's specialist row.} Our ModernBERT-base specialist, trained with the same RLCD
recipe, splits, and epoch budget as Track~B and differing only in the encoder,
measures $0.5425$ $[0.5145, 0.5685]$, below the $0.646$ the dataset card reports
(\rep{}) for its ModernBERT-base specialist; the interval excludes the reported
value. The two rows are not the same experiment. The card's row is a
frozen-encoder Adaptive Classifier (one classifier per question, mean pooling, 30
epochs; configuration published on the card and tuned on a held-out quarter of
the training split), while ours fine-tunes the full encoder with the RLCD recipe.
The gap therefore measures a training-method difference on a 149M-parameter
encoder, and no reproducibility conclusion is drawn from it (Limitations,
item~11). Under matched training the run still measures the encoder-capacity gap
cleanly, $0.23$ accuracy (paired $\Delta = -0.2245$, $p \le 2{\times}10^{-4}$)
between ModernBERT-base and the large-encoder specialists.

E3's interval discipline matters at this sample size. The case-cluster BCa interval on
\textsc{eval} winner accuracy is ${\sim}1.19\times$ wider than the naive i.i.d.\
decision-level interval (\meas{}; Figure~\ref{fig:f8}, Appendix~\ref{app:family});
decision-level resampling would have overstated precision throughout.

\section{Calibration (E4)}\label{sec:e4}

Six pre-specified methods were fitted on Track~A \textsc{calib} (200 cases, 1{,}000
decisions; and independently on Track~B's 240-case validation split), never on
\textsc{eval} or Track~B test. The six methods:
\begin{enumerate}\setlength{\itemsep}{1pt}
  \item identity, the shipped configuration including the inherited override,
  exactly as executed by \texttt{laya} 0.3.20;
  \item single temperature;
  \item per-primitive temperature (three parameters);
  \item per-option-count temperature (the shipped configuration's own
  parameterization, refitted properly);
  \item vector scaling on the \texttt{choice} head;
  \item isotonic regression (per-(primitive,\,position), one-vs-rest,
  renormalized).
\end{enumerate}
Temperatures minimize NLL on
the fitting split; the frozen selection rule picks the method with lowest fitting-split
NLL. Table~\ref{tab:t4} reports both splits; Figure~\ref{fig:f1} (\S\ref{sec:e1})
shows the identity reliability diagrams this section repairs.

\begin{table}[t]
\centering\footnotesize
\setlength{\tabcolsep}{3.5pt}
\begin{tabular}{lrr|rrrr}
\toprule
 & \multicolumn{2}{c|}{\textsc{calib} (fit split)} & \multicolumn{4}{c}{\textsc{eval} (held out, single frozen pass)} \\
Method & NLL & ECE$_{15}$ & NLL [95\% BCa] & ECE$_{15}$ [95\% BCa] & Acc. & AURC \\
\midrule
identity (shipped)      & 0.7111 & 0.2250 & 0.7027 $[0.676, 0.731]$ & 0.2035 $[0.174, 0.230]$ & 0.762 & 0.0996 \\
single temperature      & 0.5195 & 0.0228 & 0.5364 $[0.487, 0.598]$ & 0.0372 $[0.020, 0.043]$ & 0.762 & 0.0938 \\
per-primitive           & 0.5170 & 0.0226 & 0.5352 $[0.485, 0.597]$ & 0.0398 $[0.014, 0.047]$ & 0.762 & 0.0923 \\
per-option-count        & 0.5170 & 0.0226 & 0.5352 $[0.485, 0.597]$ & 0.0398 $[0.014, 0.047]$ & 0.762 & 0.0923 \\
vector (\texttt{choice})& 0.5071 & 0.0239 & 0.5346 $[0.484, 0.598]$ & 0.0314 $[0.015, 0.035]$ & 0.764 & 0.0938 \\
isotonic (\emph{frozen winner}) & \textbf{0.4463} & 0.0419 & \textbf{0.7738} $[0.626, 1.030]$ & 0.0505 $[0.028, 0.066]$ & 0.764 & 0.0993 \\
\bottomrule
\end{tabular}
\caption{\textbf{Calibration methods, Track~A (T4).} All \meas{}. Fitted on
\textsc{calib} only; \textsc{eval} touched exactly once with all methods frozen.
Accuracy is invariant under (1)--(4) by construction (temperature preserves argmax);
isotonic and vector scaling can change it and do, trivially ($+0.002$). Bold marks the
frozen selection rule's choice (lowest \textsc{calib} NLL), which has the
\emph{worst} held-out NLL in the table, worse than the uncalibrated identity. Fitted
parameters: single $T = 0.469$; per-primitive $T = (0.539, 0.431, 0.412)$ for
(\texttt{choice}, \texttt{score}, \texttt{noul}); the per-option-count method
collapses to per-primitive because the benchmark occupies only one bucket per
primitive (all \texttt{choice} at $k \le 5$). Per-method Brier and error-detection
AUROC with intervals are in \texttt{anc/metrics/e5\_e6\_eval.json}.}
\label{tab:t4}
\end{table}

\paragraph{Most of the card's miscalibration is removable by one disjointly fitted
scalar, and the shipped override is directionally backwards.} One
disjointly fitted scalar takes
held-out ECE$_{15}$ from $0.2035$ to $0.0372$ and NLL from $0.7027$ to $0.5364$
(A2/B2: $\Delta$NLL $-0.166$ $[-0.198, -0.131]$, $\Delta$ECE $-0.166$
$[-0.191, -0.107]$, both $p \le 2{\times}10^{-4}$, FDR-significant). The fitted temperature is
$T = 0.469 < 1$; the logits need \emph{sharpening}. The shipped configuration divides
logits by $T \in [1.25, 1.98]$, softening, which is the wrong direction on this
benchmark and mechanically explains the direction \S\ref{sec:underconf} measured. Per-primitive temperatures ($0.41$--$0.54$) buy almost nothing over the
single scalar ($\Delta$NLL $-0.0012$ on \textsc{eval}), and the per-option-count
parameterization, the shipped configuration's own functional form, is
\emph{identical} to per-primitive here because the benchmark's fixed schemas occupy
one option-count bucket per primitive. The shipped six-entry temperature table (of
twelve possible buckets), including the
clamped \texttt{choice:11+} entry, cannot even be exercised by the benchmark that
motivated it.

\paragraph{The frozen selection rule picked an overfit calibrator, on both tracks.}
The frozen rule (lowest fitting-split NLL) selected isotonic regression, whose
$6$--$28$ interpolation knots per (primitive,\,position) curve (median 21) won \textsc{calib} NLL
by a wide margin ($0.446$ vs.\ $0.520$) and then failed to transfer. Held-out NLL is
$0.7738$, \emph{above identity}, with a CI spanning $[0.626, 1.030]$ (the instability
signature of near-zero fitted probabilities meeting held-out one-hot targets). Its
confirmatory NLL contrast A6 does not reject ($\Delta = +0.071$
$[-0.087, +0.263]$, $p = 0.446$), one of only two failures in the 22-test executed
family. The same rule picked isotonic on Track~B's independent validation split and
its NLL contrast failed there too (C6, $p = 0.108$; \S\ref{sec:e7}), so this is a
replicated property of the selection rule at $n = 200$--$240$ fitting cases rather
than one unlucky split. Because H2 was operationalized as ``the \emph{selected} method improves
held-out NLL and ECE,'' \textbf{H2 is formally not supported on either track}, even
though every temperature-family method (A2--A5, B2--B5, C2--C5, D2--D5) rejects
decisively and isotonic's own ECE contrast (B6/D6) rejects as well. The hypothesis
failed under its frozen operationalization. A
selection-by-NLL rule at this sample size prefers expressive monotone fits that do not
generalize in NLL. We do not substitute a post-hoc winner; all downstream frozen
analyses (E5) use isotonic as selected.

This also answers \S\ref{sec:underconf}; the residual miscalibration after one
scalar is ${\sim}0.02$--$0.04$ ECE, an operational statement about repairability
rather than a unique model-versus-configuration decomposition. The
error-detection AUROC moves only from $0.766$ (identity) to $0.780$ (single
temperature) on \textsc{eval}; calibration reshapes confidence values but barely
changes their \emph{ranking}, which is what \S\ref{sec:e5} gates on.

\section{Selective escalation (E5) and cost (E6)}\label{sec:e5}

Protocol as frozen (preregistration \S0/\S5). Confidence is $q = \max_k p_k$ after the
selected calibration (isotonic, as selected; \S\ref{sec:e4}). Gating accepts a
decision when $q \ge \tau$; $\tau^{*}$ was chosen on \textsc{calib} from the
frozen twelve-point grid as the smallest value with accepted-set error
$\le 0.10$, then frozen before the single \textsc{eval} pass. On \textsc{calib} this
gave $\tau^{*}_{\mathrm{dec}} = 0.70$ (coverage $0.644$, accepted error $0.0916$) and
$\tau^{*}_{\mathrm{case}} = 0.80$ (a case escalates if \emph{any} of its five
decisions falls below $\tau$; coverage on \textsc{calib} already only $0.050$).
Because no runnable escalation target exists in this study (\S\ref{sec:comparators}),
end-to-end accuracy is reported \emph{only} as the symbolic form
$c \cdot a + (1-c)\,a_T$ and its oracle bound ($a_T = 1$); E6 cost likewise as
$c_L + (1-c)\,c_T$ with $c_T$ symbolic.

\paragraph{Operating point on {\normalfont\textsc{eval}} (all {\normalfont\meas{}}, single
frozen pass, 200 cases / 1{,}000 decisions).} Decision-level gating at $\tau^{*} = 0.70$: coverage
$0.672$ $[0.637, 0.704]$, accepted-set accuracy $0.8601$ $[0.831, 0.886]$,
oracle-bounded end-to-end accuracy $0.906$ $[0.885, 0.923]$, escalation rate $0.328$.
\textbf{The frozen gate misses its target out-of-sample.} Accepted-set error is
$0.1399$ against the frozen $0.10$; the \textsc{calib}-feasible threshold
did not transfer ($0.0916 \to 0.1399$), a $53\%$ relative degradation. H3(a) therefore
fails on Track~A (and, independently, on Track~B: $0.0866 \to 0.1028$;
\S\ref{sec:e7}). H3(b), that the confidence signal beats random escalation at matched
rate, holds decisively: E1t $\Delta = +0.0646$ $[0.0502, 0.0785]$, one-sided
$p \le 10^{-4}$, FDR-significant. So the signal is real but the frozen threshold's error
guarantee is not; a deployment would need to re-validate $\tau$ on fresh data rather
than trust a 200-case fit. Accuracy at fixed coverage (winner confidence ranking):
$0.800$ at $90\%$, $0.839$ at $75\%$, $0.908$ at $50\%$, $0.976$ at $25\%$, from
$0.764$ at full coverage; AURC $0.0993$ (winner) vs.\ $0.0996$ (identity),
consistent with \S\ref{sec:e4}; calibration barely changes the ranking.

\paragraph{Why the frozen gate missed (exploratory, post hoc).}
A decomposition of the stored per-decision records (deviation D9,
Appendix~\ref{app:deviations}; artifact \texttt{e11\_tau\_decomposition.json})
shows the miss was built in at fitting time. Under the identity ranking, which no
calibration map fitted on \textsc{calib} can change, accepted-set error at matched
coverage $0.644$ is $0.143$ on \textsc{calib} against $0.127$ on \textsc{eval}, so
the held-out half is not harder. Under the isotonic ranking the same comparison
reads $0.092$ against $0.130$, an in-sample optimism of $0.051$ on \textsc{calib}.
The mechanism is that the calibrator and the threshold were fitted on the same
1{,}000 decisions, and isotonic's in-sample fit sinks its own training errors below
the gate. The one-parameter single-temperature calibrator, which cannot re-rank
decisions within a primitive, lands at the target out of sample ($\tau = 0.75$:
\textsc{calib} $0.094$, \textsc{eval} $0.100$ at coverage $0.598$). A counterfactual
rule that requires the 95\% upper confidence bound on \textsc{calib} accepted error
to clear the target picks $\tau = 0.80$ under both a Wilson bound ($0.086$) and a
case-cluster bootstrap 95th percentile ($0.082$), delivering $0.101$ at coverage
$0.534$ on \textsc{eval}. The frozen point-estimate rule's $\tau^{*} = 0.70$ had a
\textsc{calib} Wilson upper bound of $0.116$, already above the target it was meant
to guarantee.

\paragraph{Case-level gating collapses.} At $\tau^{*}_{\mathrm{case}} = 0.80$,
requiring all five decisions confident leaves coverage $0.055$ $[0.025, 0.085]$
(eleven accepted cases), with accepted-set accuracy $0.8545$ (error $0.145$, also
missing target) and no significant advantage over random case escalation (E1t case
variant: $\Delta = +0.005$, $p = 0.075$; secondary, outside the confirmatory family).
The all-five-confident requirement is too conservative to be useful at these
confidence levels; decision-level gating is the only viable mode measured here.

\begin{figure}[t]
\centering
\includegraphics[width=0.85\linewidth]{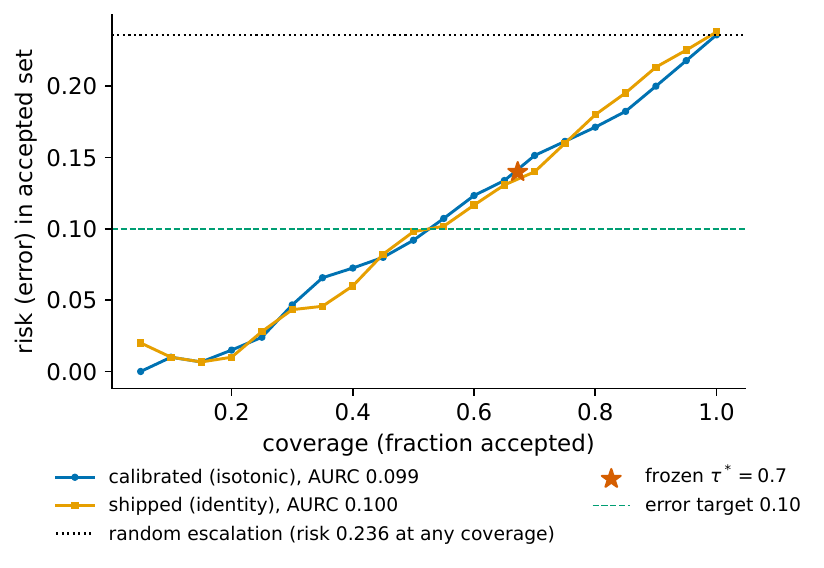}
\caption{\textbf{F2} (\meas{}). Risk--coverage on Track~A \textsc{eval}, shipped
identity vs.\ frozen winner confidence, random-escalation reference line, the frozen
$\tau^{*}$ operating point (star), and the frozen $0.10$ error target (dashed).
The operating point sits visibly above the target line.}
\label{fig:f2}
\end{figure}

\begin{figure}[t]
\centering
\includegraphics[width=0.72\linewidth]{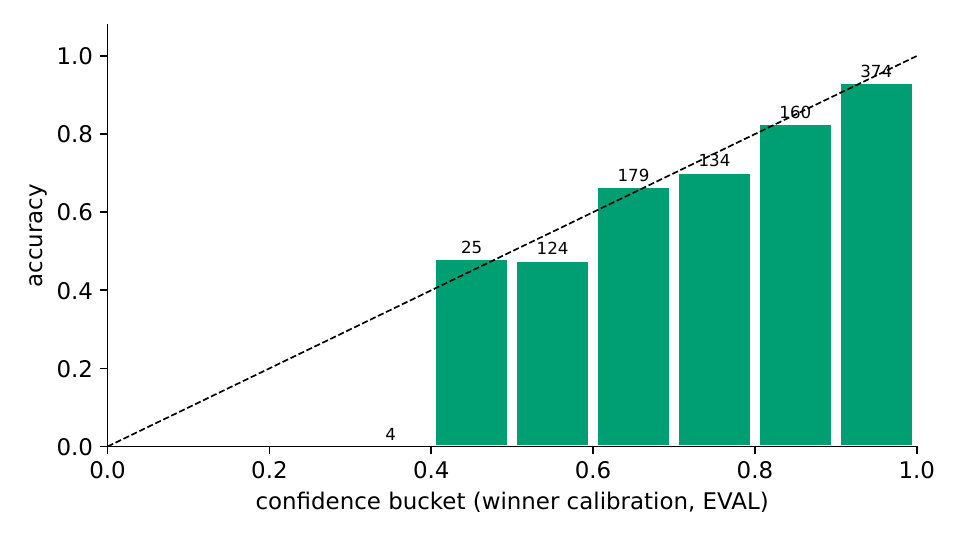}
\caption{\textbf{F3} (\meas{}). Accuracy by confidence bucket on \textsc{eval}
with case-clustered intervals. Ordering is monotone above $q \ge 0.5$ except a flat
step at $[0.5,0.6)$ vs.\ $[0.4,0.5)$ ($0.476$ vs.\ $0.480$, $n = 124$ vs.\ $25$).}
\label{fig:f3}
\end{figure}

\begin{figure}[t]
\centering
\includegraphics[width=\linewidth]{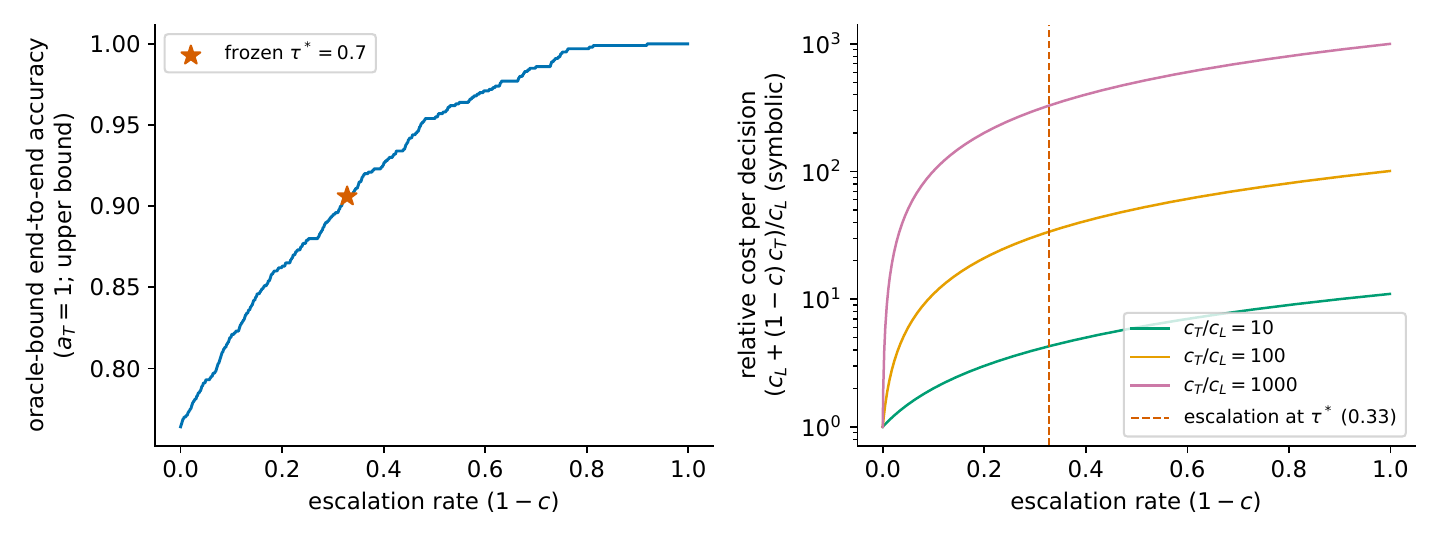}
\caption{\textbf{F4} (\meas{}/symbolic). Oracle-bounded end-to-end accuracy and the
symbolic cost form $c_L + (1-c)\,c_T$ across the $\tau$ grid; the bound is labeled a
bound.}
\label{fig:f4}
\end{figure}

\paragraph{High-confidence errors in security and invoice decisions.} Among
\emph{accepted} decisions at $\tau^{*} = 0.70$ on \textsc{eval},
\texttt{security\_incidents} has $26$ errors in $177$ accepted decisions
($14.7\%$) and \texttt{invoice\_processing} $16/187$ ($8.6\%$). The single most
confident security error is instructive: case \texttt{security\_incidents\_000043},
severity (\texttt{score}), predicted $3$ at calibrated confidence $1.0$ against a
\emph{unanimous} teacher gold of $4$ (\texttt{label\_agreement} records no split,
ruling out a teacher-ambiguity artifact). The top invoice errors are mostly
teacher-split cases, ambiguous items the gate should have escalated. A cascade whose accepted set contains
confidence-$1.0$ security misclassifications is not made safe by an $86\%$ average.
Isotonic's saturating upper knots (which map high raw confidence to exactly $1.0$)
make the top of the confidence scale uninformative, an argument for the temperature
family on operational grounds even before A6's statistical failure.

\paragraph{Cost and latency (E6).} Latency and throughput are the canonical E1
measurements of \S\ref{sec:e1} (this hardware only; a step-8 re-measurement that ran
concurrently with Track~B training measured p50 $204$~ms and is superseded as
contended, Appendix~\ref{app:deviations}).
Cascade cost per decision is $c_L + (1-c)\,c_T$ with measured
$c_L \approx 18.4$~ms/decision amortized (batch~1, $2{,}000$ decisions in $36.8$~s)
and $c_T$ symbolic; at the frozen operating point the escalation term is
$0.328\,c_T$. Third-party cascade economics in adjacent settings appear in
\S\ref{sec:related}; no comparison to them is claimed.

\section{Robustness (E7), ablations (E8), and the Track~B replication}\label{sec:e7}

All E7 perturbations ran on \textsc{calib} cases only (no \textsc{eval} exposure;
\textsc{calib} is itself half of the official test split, per the exposure ledger in
\S\ref{sec:prereg}); everything in this section is exploratory under the frozen plan except the
Track~B family rows (C/D/E2t), which are confirmatory. All numbers \meas{}.

\paragraph{Option-order permutations.} Across four seeded permutations of every
\texttt{choice} decision's option order ($300$ decisions $\times$ 4 runs), $6.75\%$ of
predictions flip; accuracy ranges $0.727$--$0.763$ across runs against $0.747$
original. The predicted-position histogram is flat over positions $0$--$3$
($287/291/243/293$) with position~4 at $86$, but position~4 exists only for $k=5$
questions, so the small count reflects option-count composition rather than avoidance. This
is far milder than the option-position pathologies documented for generative LLM
judges \citep{zheng2024mcq,pezeshkpour2024,wang2024fair}; a typed encoder head appears structurally less
order-sensitive, though $6.75\%$ of flips is not zero and would touch
${\sim}20$ decisions per 400-case batch.

\paragraph{Surface perturbations (substitute for paraphrase; deviation D2).} The
planned paraphrase test named no runnable paraphraser, so two deterministic
surface transforms substitute, and are labeled as such. Lowercasing with punctuation
stripped flips $1.7\%$ of predictions (accuracy $0.772 \to 0.770$); a neutral prefix
flips $1.4\%$ ($\to 0.777$). Mean confidence shifts are ${<}0.003$. This tests
format brittleness only, \emph{not} semantic robustness; no paraphrase claim is
made.

\begin{figure}[t]
\centering
\includegraphics[width=0.85\linewidth]{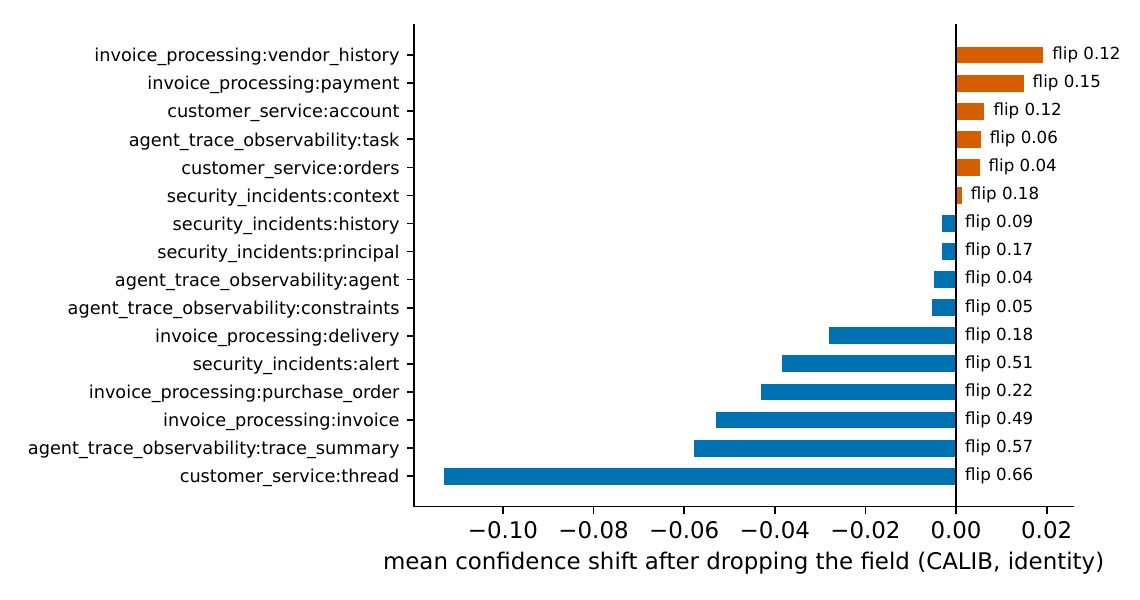}
\caption{\textbf{F7} (\meas{}). Per-field flip rates and
mean confidence shifts under field removal; load-bearing fields (annotated) reduce
confidence, the correct direction for a cascade.}
\label{fig:f7}
\end{figure}

\paragraph{Missing evidence lowers confidence.}\begin{sloppypar}\hbadness=10000 Dropping a
load-bearing state field before re-scoring changes $49$--$66\%$ of that workflow's
predictions and cuts its accuracy roughly in half (e.g.\
\texttt{customer\_service:thread}: flip rate $0.656$, accuracy $0.800 \to 0.332$;
\texttt{security\_incidents:alert}: $0.508$, $\to 0.500$;
\texttt{invoice\_processing:invoice}: $0.492$, $\to 0.476$;
\texttt{agent\_trace\_observability:trace\_summary}: $0.572$, $\to 0.324$). The
assessor-relevant finding is the confidence direction. For every load-bearing field
the mean confidence \emph{drops} (e.g.\ $-0.113$ for \texttt{thread}, $-0.058$ for
\texttt{trace\_summary}), i.e.\ the model becomes less confident when evidence is
missing, which is the gate behavior a cascade needs.
Peripheral fields (\texttt{orders}, \texttt{constraints}) flip ${\le}5\%$ with
negligible shifts. Figure~\ref{fig:f7} shows all fields.\end{sloppypar}

\paragraph{Label shift and option-count stress.} Importance-reweighting \textsc{calib}
to a uniform gold-label distribution per (workflow,\,question) drops accuracy from
$0.772$ to $0.688$; roughly $0.08$ of headline accuracy rides on the benchmark's
skewed label priors. Synthetic single-question retrieval probes at
$k \in \{5,10,20,30\}$ options all score $100\%$ ($40$ probes each; trivial by design,
not comparable to benchmark accuracy). They confirm the \emph{only} live site of the
shipped \texttt{choice:11+} clamp; at $k \ge 11$ the raw temperature $0.1006$ is
clamped to $0.5$, and mean probe confidence jumps to $0.97$/$0.94$. The clamp
region is exactly where the configuration sharpens most aggressively, on the least
tested inputs.

\begin{figure}[t]
\centering
\includegraphics[width=\linewidth]{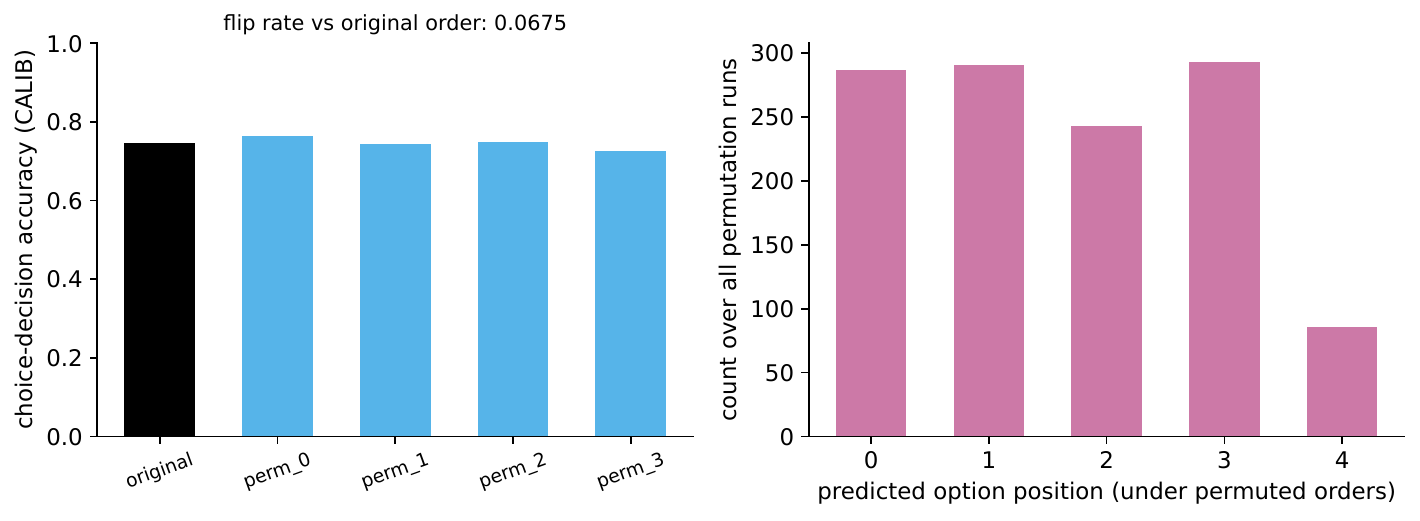}
\caption{\textbf{F6} (\meas{}). Accuracy by permutation run and
predicted-position histogram.}
\label{fig:f6}
\end{figure}

\paragraph{Ablations (E8; stored {\normalfont\textsc{eval}} records only, no new forwards).} The
calibration $\times$ cascade $2 \times 2$ shows the interaction that matters. Gating
\emph{shipped} confidences at $\tau = 0.70$ accepts only $18.8\%$ of decisions (at
$0.995$ accepted accuracy) because under-confident scores rarely clear the bar, while
the calibrated pipeline accepts $67.2\%$ at $0.860$; calibration is what makes the
gate's coverage usable at all. Among confidence statistics at matched $67.2\%$
coverage, negative entropy edges out max-probability ($0.8661$ vs.\ $0.8601$ accepted
accuracy) and margin trails ($0.8542$); ordinal variance on \texttt{score} decisions
only is worst ($0.8290$). Per primitive, isotonic repairs ECE$_{15}$ from
$0.238/0.193/0.186$ to $0.066/0.035/0.063$ (\texttt{choice}/\texttt{noul}/\texttt{score})
with accuracy essentially unchanged. The naive-vs-cluster interval contrast is
case-cluster BCa width $0.064$ vs.\ naive i.i.d.\ $0.054$, ratio $1.185$
(Figure~\ref{fig:f8}).

\subsection{The Track~B clean replication and the price of contamination}\label{sec:trackb}

Track~B retrains from the pinned \texttt{laya} base with the card's RLCD recipe on
case-level splits ($960$ fine-tune / $240$ validation cases, seed $20260926$; one
Apple-silicon device), then touches the 400-case test split exactly once with its own
independently frozen pipeline (validation-fitted methods; winner again isotonic;
$\tau^{*}_{\mathrm{dec}} = 0.75$). The identity configuration is the trained
checkpoint's own per-type temperatures, fitted on the 240 validation cases during
training ($1.145$, $1.122$, $1.197$ for \texttt{choice}, \texttt{score},
\texttt{noul}; run manifest), and methods (2)--(6) are fitted on top of it.

\textbf{Replication.} Test accuracy $0.7710$ $[0.7495, 0.7905]$ vs.\ the released
checkpoint's $0.7670$; paired case-level difference $+0.0040$
$[-0.009, +0.017]$, $p = 0.568$; the card's training recipe replicates within noise
under a cleaner protocol, on different hardware, with case-level splits. But the
\emph{identity} configurations differ sharply in probability quality. Track~B's
validation-fitted temperatures give test NLL $0.6271$ vs.\ $0.7069$ and ECE$_{15}$
$0.1509$ vs.\ $0.2142$ (paired $\Delta$NLL $-0.0798$ $[-0.089, -0.070]$,
$\Delta$ECE $-0.0634$ $[-0.075, -0.048]$, both $p \le 2{\times}10^{-4}$). Same recipe,
same accuracy, but this cross-checkpoint contrast alone cannot isolate the cause;
Track~B also carries newly trained weights, a different optimization trajectory, and
different hardware. The calibration-protocol attribution rests on the within-Track-B
contamination probe below, which holds the weights fixed.

\textbf{The selection-rule failure replicates.} Track~B's frozen rule again picked
isotonic (validation NLL $0.489$ vs.\ $0.560$ single-temperature), and again the
held-out NLL contrast failed (C6: $\Delta = -0.046$ $[-0.094, +0.012]$, $p = 0.108$)
while every temperature-family contrast (C2--C5, D2--D6) rejected at
$p \le 2{\times}10^{-4}$.
Track~B's cascade also repeats Track~A's pattern. The frozen
$\tau^{*}_{\mathrm{dec}} = 0.75$ (validation accepted error $0.0866$) misses the
target on test ($0.1028$; coverage $0.5835$ $[0.557, 0.609]$, accepted accuracy
$0.8972$ $[0.876, 0.914]$), and beats random escalation decisively (E2t:
$\Delta = +0.0757$ $[0.0661, 0.0851]$, one-sided $p \le 10^{-4}$). The two tracks,
with independent fitting splits, recorded the same two qualitative failures, which
places both in the frozen procedure at this sample size rather than in either
split. The \S\ref{sec:e5} decomposition holds in the same direction here
(validation $0.0866 \to$ test $0.1028$; single-temperature gating on test reaches
$0.105$ at $\tau = 0.70$; deviation D9).

\textbf{Contamination probe.} Refitting per-primitive temperatures on $400$ decisions
the model \emph{trained on} (the released checkpoint's own protocol, per the
fine-tuning notebook linked from the model card) yields temperatures
$(0.400, 0.309, 0.219)$ vs.\ the clean
validation's $(0.582, 0.541, 0.638)$; training-set confidence looks sharper than it
is, so the fit sharpens beyond what held-out data supports. Held-out test ECE$_{15}$ degrades from $0.0236$ (clean fit) to
$0.0873$ (contaminated fit), a $3.7\times$ inflation at identical accuracy. This is an
upper-bound probe (the notebook's actual holdout was item-level within partially seen
cases, between these regimes), but it measures the mechanism and its direction
directly.

\section{External validity (E9)}\label{sec:e9}

\descopedmark

The pre-freeze experiment plan (not released; summarized in the preregistration's
\S0.a) specified an external set of 120 independently sourced, human-adjudicated
cases (two blinded annotators, third-adjudicator protocol, chance-corrected agreement
reporting). The preregistration's \S0.a amendment records that no annotator pool
exists for this version and exercises the plan's stated fallback (its \S3.3, option
3): state so ``plainly'' and restrict every conclusion to the synthetic benchmark
(the word is the plan's).
We do that; no numbers from the planned external set exist, and the
two external-validity members of the confirmatory family (F1t/F2t; hypothesis H4)
are reported as not executed (Appendix~\ref{app:family}). What follows in
\S\ref{sec:e10} is a separate, post-registration exploratory probe and is no
substitute for that set.

\subsection{An exploratory out-of-distribution probe (E10)}\label{sec:e10}

Because the planned external set is descoped, we ran one \emph{post-registration
exploratory} probe of our own, labeled as such on every artifact and never entering
the confirmatory family (H4 remains untested; F1t/F2t remain not executed). The probe
asks whether the released checkpoint (F16 weights) transfers zero-shot to evidence-grounded
claim verification over documents from its own four workflow domains. We authored a
corpus of 24 fictional workflow documents (six per domain, 150--192 words each) and
120 typed claims about them, balanced 60 supported / 60 not-supported across six
claim types (paraphrase, inference, number, entity, contradiction, unverifiable; 20
each), with verbatim evidence spans recorded for every verifiable claim and a
validator script enforcing the corpus constraints. The harness is standard
retrieval-augmented \texttt{noul} querying (sentence-aware ${\sim}55$-word chunks,
MiniLM-L6-v2 embeddings, cosine top-$k{=}5$, one yes/no question per claim through
\texttt{predict\_batch}), with document-level cluster bootstrap intervals. All
\meas{}, exploratory.

The checkpoint does not transfer. Accuracy is $0.617$ $[0.533, 0.683]$ (balanced
accuracy the same), against $0.858$ (\meas{}) on the benchmark's own in-distribution
\texttt{noul} decisions, a drop of roughly $0.24$. The two
\texttt{noul} tasks are related but not identical, so the delta is a directional
signal rather than a like-for-like measurement. One in three unsupported claims is accepted (false
acceptance $0.333$ $[0.218, 0.439]$). Confidence barely ranks the errors: AUROC
$0.589$ $[0.495, 0.683]$, and the confidence distribution compresses below
${\sim}0.77$, so the frozen gating rule is \emph{infeasible} on this set. No
grid threshold reaches the $0.10$ error target at nonzero coverage, and coverage
dies before error improves (at $\tau = 0.65$: coverage $0.175$, accepted error still
$0.238$). Per claim type, contradiction ($0.750$) and number ($0.700$) are detected best;
entity swaps and one-step inference sit at chance ($0.500$ each), the failures most
dangerous for evidence checking. Retrieval is not the whole explanation (evidence
recall@5 is $0.84$).

On this probe's evidence, the in-distribution results of
\S\ref{sec:e2}--\ref{sec:e7} are properties of the
benchmark match, and nothing in this paper licenses deploying this checkpoint as a
claim verifier or on any distribution it was not fine-tuned for. This probe is our
own design and instrumentation end to end (corpus, claims, validator, harness); its
corpus, per-decision predictions, validation record, and results are in the
ancillary files (\texttt{anc/e10/}).

\section{Discussion}\label{sec:discussion}

\paragraph{Benchmark circularity, and what $0.767$ can and cannot mean.}
The measured accuracy ($0.767$; \meas{}) sits \emph{above} the teacher's own
self-agreement ceiling ($0.735$; \rep{}) and well above the perfect-factor-knowledge
reference ($0.704$; \rep{}). The dataset card's own interpretation rule says scores
much above ${\sim}0.75$ indicate a model has learned the teacher's idiosyncrasies
rather than the task. The correct reading of E1 is therefore that a
421M specialist fine-tuned on these four workflows saturates agreement with a noisy
synthetic teacher on held-out cases from the same generator; no ``beats the
teacher'' reading survives that ceiling. Teacher agreement is not
correctness \citep{gudibande2023}; a model can be penalized for being right where
the teacher is wrong.
This is also why no superiority claim over JEV is available from any table here. The
third-party JEV row ($0.727$; \tp{}) was produced by a different party under different
prompts and pipeline on a benchmark whose ceiling is $0.735$. The specialist/generalist
grouping makes the rows non-comparable by the dataset card's own discipline.
A zero-shot hosted model on the same card also clears the ceiling ($0.768$;
\tp{}), so the ceiling is a reading rule for a specialist fitted to this teacher
rather than a hard cap on any model; the claim here concerns only what the
specialist's score can mean.

\paragraph{Direction of miscalibration is an operational quantity.}
Two checkpoints with ECE $0.21$ can fail in
opposite directions. An over-confident assessor silently passes errors through a
cascade gate, while an under-confident one escalates too much and erodes the cost
case. The shipped Laya artifact is the second kind (\S\ref{sec:underconf}), with the
mechanism measured in \S\ref{sec:e4}. After proper calibration the signed gap flips
to slightly \emph{positive} ($+0.028$ for the single temperature and $+0.033$ for
the selected isotonic map on \textsc{eval}), mild over-confidence, which
is the regime in which the E5 gate under-delivers. Direction of
miscalibration, not its magnitude, determines which cascade failure mode a deployment
gets, and the threshold was chosen on the calibrator's own fitting data, so its
in-sample accepted-set error understated the held-out error (the \S\ref{sec:e5}
decomposition).

\paragraph{Procedure-level failures the preregistration recorded.}
One frozen design choice produced both recorded failures. The calibrator and the
threshold were fitted on the same split, under an NLL-based selection rule that
favors expressive calibrators, so the selected isotonic map overfit (A6/C6). The
same in-sample optimism let $\tau^{*}$ clear a target it could not hold out of
sample (H3a; decomposition in \S\ref{sec:e5}). Without preregistration both
would have been silently ``fixed'' (swap in the vector method, nudge $\tau$ to
$0.75$) and the paper would have reported only successes. The recorded failures
measure the generalization gap of the \emph{procedure}, which is what a
practitioner reusing this recipe on their own
200-case calibration set needs to know \citep{nosek2018}. The reusable lesson is to
cross-fit (fit the calibrator on one half of the fitting split, choose $\tau$ on
the other) or to select $\tau$ by an upper confidence bound on the fitting-split
error.

\paragraph{Where a compact typed assessor is the right tool.}
The measured operating point, ${\sim}19$\,ms per decision at batch~1 on a
laptop-class accelerator with full distribution outputs (\meas{}; \S\ref{sec:e1}),
is the argument for the interface. It is local, auditable, and cheap enough to run on
every decision, with typed label spaces fixed by the workflow. Published cascades in adjacent settings
\citep{li2026,gisserot2026,chen2024frugalgpt,jitkrittum2023} suggest such a first
stage can carry most of a workload \emph{when its confidence orders its errors}, and
that thresholds must be validated locally. This study now has the measured version of that advice. The
confidence signal beat random escalation at matched rate on both tracks, and
calibration is what made the gate's coverage usable at all (E8).
In-distribution, the measured
posture is that a compact typed assessor plus one fitted temperature is a useful
triage stage at roughly two-thirds coverage. That holds provided the error target is
treated as an estimate to monitor, not a guarantee, and the accepted set is audited
(\S\ref{sec:e5}). Out of distribution, the exploratory probe measured no transfer
(\S\ref{sec:e10}); nothing in this paper counters that.

\section{Limitations}\label{sec:limitations}

\begin{enumerate}
  \item \textbf{Single checkpoint, single configuration.} One released artifact at one
  pinned revision, as shipped; no model-family sweep, no seed variation.
  \item \textbf{Synthetic, teacher-labeled benchmark.} Gold is the mean of three
  samples from one ${\sim}$4B-class teacher; every metric is agreement with that
  teacher rather than correctness \citep{gudibande2023}; cases are template-generated,
with no fielded deployment behind them.
  \item \textbf{No compute-matched architectural comparison.} Comparators differ in
  size, training exposure, and interface; grouping by task access mitigates but does
  not remove this.
  \item \textbf{No TypeSafe JEV access.} All JEV figures are third-party on different
  pipelines and samples; nothing here supports a head-to-head ranking in either
  direction.
  \item \textbf{No preregistered external validity set in this version} (descoped;
  \S\ref{sec:e9}); every confirmatory conclusion is restricted to the synthetic
  benchmark, and H4 goes untested. The E10 probe (\S\ref{sec:e10}) is exploratory,
  post-registration, self-authored (120 claims, 24 documents), and its claims are
  correspondingly weak, a transfer-failure signal rather than a measurement of
  external validity.
  \item \textbf{No independent annotator pool.} The external-set annotation protocol
  (two blinded annotators, adjudication, agreement statistics) exists only in the
  unreleased pre-freeze plan; no human adjudication of any model output occurred
  in this version.
  \item \textbf{Single hardware configuration.} Apple M5~Max / MPS; latency and
  throughput are not comparable to the card's T4 numbers or to any served deployment,
  and the observed non-monotonic batching behavior is unexplained.
  \item \textbf{Track A's fitting data is structurally contaminated.} The released
  checkpoint saw all 1{,}200 training cases, so Track~A's \textsc{calib} is carved
  from the test split as mitigation; only Track~B is clean end-to-end.
  \item \textbf{Small fitting and evaluation splits.} Track~A fits on 200 cases and
  evaluates on 200; both replicated frozen-procedure failures (isotonic selection,
  $\tau^{*}$ transfer) are properties \emph{at this sample size} and may vanish with
  more fitting data. Conversely, the $\tau$ miss means the study cannot certify any
  error target at $n = 200$, and the exploratory decomposition in \S\ref{sec:e5}
  traces that miss to in-sample optimism of the same-split fit (deviation D9).
  \item \textbf{Soft accuracy did not reproduce.} It measures
  $0.509$ against a reported $0.471$ with a CI excluding the reported value; an
  fp32/CPU rerun rules out precision and backend (deviation D10), a version
  bisection rules out every released harness version from 0.3.3 to 0.3.20
  (deviation D11), and what remains, the card-side evaluation pipeline or data
  state, is not recoverable from the card (\S\ref{sec:e1}).
  \item \textbf{The ModernBERT-base comparator uses a different recipe from the
  card's row.} Our RLCD-trained specialist measures $0.5425$ against the dataset
  card's frozen-encoder Adaptive-Classifier specialist at $0.646$; the training
  methods differ, so the row is a recipe comparison, not a reproduction of the
  card's experiment (\S\ref{sec:e2}).
  \item \textbf{Shipped confidence carries a known configuration defect} (a
  training-slice temperature fit, \rep{} on the model card, overridden by an
  inherited table whose out-of-range entry \texttt{laya} 0.3.20 clamps; \meas{}),
  the card says to treat it as uncalibrated, and \texttt{choice}-primitive
  ECE ($0.255$) is worse than the aggregate ($0.214$); the headline calibration
  number flatters the weakest primitive.
  \item \textbf{E1 predates the preregistration} (exempt because it fits nothing),
  so H1's numerical criterion is a post-hoc reporting structure, and the E1-derived
  under-confidence observation informed the framing of the frozen E4 questions.
  \item \textbf{The paraphrase test was substituted} (deviation D2); the surface
  transforms that ran test format brittleness only, and semantic paraphrase robustness
  is unmeasured.
  \item \textbf{E7's option-count probes are synthetic retrieval tasks}, trivial by
  construction; they locate where the shipped clamp is live but say nothing about accuracy
  at high option counts on real questions.
  \item \textbf{Latency/throughput are single-machine numbers.} One planned
  re-measurement was invalidated by resource contention with concurrent training and
  is superseded by the uncontended E1 measurement (Appendix~\ref{app:deviations}).
\end{enumerate}

\section{Ethics and responsible use}\label{sec:ethics}

The four workflows include payment release and security containment. Nothing in this
paper supports using this checkpoint, or any assessor, as the sole arbiter of
such decisions. The confidence signal is measurably miscalibrated as shipped, the
calibrated cascade still missed its frozen error target on both tracks, and the
accepted set contained high-confidence security errors (\S\ref{sec:e5}). The
defensible deployment posture remains decision support with human accountability,
thresholds validated locally on the deployment's own distribution, and periodic
re-checks. The benchmark is synthetic (no personal data,
no human subjects); the planned external-set collection, when it happens, will
carry its own annotation-ethics protocol. The dataset is Apache-2.0; model and
code artifacts are used under their published terms; no credentials or account
metadata appear in the released artifacts.

\section{Independence statement}\label{sec:coi}

The author has no affiliation with, funding from, or financial interest in
Convai Innovations (publisher of the model, its base model, and the
\texttt{laya} package), the LocalLLaMA Hugging Face organization (publisher of
the benchmark), or TypeSafe. E1 is an independent reproduction. The study's
integrity measures are: the analysis plan was frozen before any fitting touched
the test split (\S\ref{sec:prereg}); the acceptance gate's one timing failure is
disclosed in \texttt{gate.json}; results unfavorable and favorable to the
artifact are reported with the same prominence; per-decision artifacts for every
headline run are released so any party can recompute every number.

\section{Reproducibility statement}\label{sec:repro}

All artifacts are pinned by immutable revision, and every run writes a JSON manifest
(source-tree content hash, HF revision SHAs, package versions, hardware, batch size,
seed, wall-clock, exact CLI invocation). Pins:
{\raggedright
\begin{itemize}\setlength{\itemsep}{1pt}
\item model \texttt{convaiinnovations/laya-typed-decisions} at revision
  \texttt{1a793eb568e6718f15941d08f85432581df534e3}
  (weights SHA-256 \texttt{4fa56de7\ldots});
\item dataset \texttt{LocalLLaMA/typed-decisions} at revision
  \texttt{f7a2487edd7a043a5441a5e9ccc7fe5ddbd9ebe8};
\item base model \texttt{convaiinnovations/laya} at revision \texttt{55cf4c4e\ldots};
\item harness \texttt{laya} 0.3.20, Python 3.12.14, torch 2.14.0,
  transformers 5.17.0, \texttt{USE\_TF=0}; Apple M5~Max, 64\,GB, \texttt{mps}.
\end{itemize}}
The analysis plan was frozen on 2026-09-26 at 15:50~UTC, before any fitting touched
the test split; the frozen document is released verbatim as an ancillary file and
its SHA-256 is listed in \texttt{anc/README.txt}. Timestamps are taken from the run
manifests, which are also released. The canonical E1 artifacts (the metrics JSON,
the 2{,}000-row per-decision prediction file, and the run manifest
\texttt{20260926T120855Z}) are in the ancillary files, alongside
\texttt{gate.json}, the pass/fail record of the preregistration's acceptance gate
(its prereg-timing item is recorded as a disclosed failure; \S\ref{sec:prereg}),
and \texttt{anc/README.txt}, which lists the SHA-256 digest of every ancillary
file. The Track~A split
file's SHA-256 (\texttt{7719b147\ldots}) is recorded in the preregistration; cluster
bootstrap uses $n_{\mathrm{boot}} = 10{,}000$, seed 0; Track~A/B split seed 20260926;
Track~B trains from the pinned \texttt{convaiinnovations/laya} base, and the
ModernBERT-base specialist pins \texttt{answerdotai/ModernBERT-base} at revision
\texttt{8949b909\ldots} (all in \texttt{pins.json}).
In the ancillary files: per-decision predictions for every run (released full test,
both Track~A halves, Track~B test, both zero-shot comparators), \texttt{pins.json}
with checksums, run manifests, the metric suite with its unit tests, and the frozen
preregistration document.

\section*{Acknowledgements and AI-assistance disclosure}

This manuscript and its evaluation harness were prepared with substantial assistance
from large language models, used for literature synthesis, drafting, and code; the
study design constraints, integrity rules, and authorization decisions are the
author's. The experiments were executed by LLM agents on 2026-09-26, with
exploratory reruns on 2026-09-27, under the author's direction; the author reviewed every reported number
against the released artifacts. The two internal adversarial reviews of
Appendix~\ref{sec:timeline} were produced by two LLM systems; their
full reports are retained privately by the author.
Every measured number originates from a logged run whose manifest is in the
ancillary files; every reported or third-party number is attributed to its
public source with access date; and every bibliography entry was verified against its
primary record (arXiv or publisher metadata) on 2026-09-26 before inclusion. The
author reviewed the manuscript and takes responsibility for its content. This
disclosure is retained in view of arXiv's policy on undisclosed
LLM-generated content and hallucinated references. Appendix~\ref{sec:timeline}
records how the study was executed, including the role of LLM agents and the
internal adversarial reviews that preceded this version.

\bibliographystyle{plainnat}
\bibliography{refs}

\begin{thebibliography}{23}
\providecommand{\natexlab}[1]{#1}
\providecommand{\url}[1]{\texttt{#1}}
\expandafter\ifx\csname urlstyle\endcsname\relax
  \providecommand{\doi}[1]{doi: #1}\else
  \providecommand{\doi}{doi: \begingroup \urlstyle{rm}\Url}\fi

\bibitem[Benjamini and Hochberg(1995)]{benjamini1995}
Yoav Benjamini and Yosef Hochberg.
\newblock Controlling the false discovery rate: A practical and powerful approach to multiple testing.
\newblock \emph{Journal of the Royal Statistical Society: Series B}, 57\penalty0 (1):\penalty0 289--300, 1995.

\bibitem[Boizard et~al.(2025)Boizard, Gisserot-Boukhlef, Alves, Martins, Hammal, Corro, Hudelot, Malherbe, Malaboeuf, Jourdan, Hautreux, Alves, El~Haddad, Faysse, Peyrard, Guerreiro, Fernandes, Rei, and Colombo]{boizard2025}
Nicolas Boizard, Hippolyte Gisserot-Boukhlef, Duarte~M. Alves, Andr{\'e} Martins, Ayoub Hammal, Caio Corro, C{\'e}line Hudelot, Emmanuel Malherbe, Etienne Malaboeuf, Fanny Jourdan, Gabriel Hautreux, Jo{\~a}o Alves, Kevin El~Haddad, Manuel Faysse, Maxime Peyrard, Nuno~M. Guerreiro, Patrick Fernandes, Ricardo Rei, and Pierre Colombo.
\newblock {EuroBERT}: Scaling multilingual encoders for {European} languages, 2025.
\newblock arXiv:2503.05500.

\bibitem[Chen et~al.(2024)Chen, Zaharia, and Zou]{chen2024frugalgpt}
Lingjiao Chen, Matei Zaharia, and James Zou.
\newblock {FrugalGPT}: How to use large language models while reducing cost and improving performance.
\newblock \emph{Transactions on Machine Learning Research}, 2024.

\bibitem[Chow(1970)]{chow1970}
C.~K. Chow.
\newblock On optimum recognition error and reject tradeoff.
\newblock \emph{IEEE Transactions on Information Theory}, 16\penalty0 (1):\penalty0 41--46, 1970.

\bibitem[Davison and Hinkley(1997)]{davison1997}
Anthony~C. Davison and David~V. Hinkley.
\newblock \emph{Bootstrap Methods and their Application}.
\newblock Cambridge University Press, 1997.

\bibitem[Efron and Tibshirani(1993)]{efron1993}
Bradley Efron and Robert~J. Tibshirani.
\newblock \emph{An Introduction to the Bootstrap}.
\newblock Chapman \& Hall, 1993.

\bibitem[Geifman and El-Yaniv(2017)]{geifman2017}
Yonatan Geifman and Ran El-Yaniv.
\newblock Selective classification for deep neural networks.
\newblock In \emph{Advances in Neural Information Processing Systems 30}, 2017.

\bibitem[Gisserot-Boukhlef et~al.(2026)Gisserot-Boukhlef, Boizard, Malherbe, Hudelot, and Colombo]{gisserot2026}
Hippolyte Gisserot-Boukhlef, Nicolas Boizard, Emmanuel Malherbe, C{\'e}line Hudelot, and Pierre Colombo.
\newblock {BERT}-as-a-{Judge}: A robust alternative to lexical methods for efficient reference-based {LLM} evaluation.
\newblock In \emph{Proceedings of the Conference on Language Modeling (COLM)}, 2026.
\newblock Camera-ready (arXiv:2604.09497v2, 20 July 2026).

\bibitem[Gneiting and Raftery(2007)]{gneiting2007}
Tilmann Gneiting and Adrian~E. Raftery.
\newblock Strictly proper scoring rules, prediction, and estimation.
\newblock \emph{Journal of the American Statistical Association}, 102\penalty0 (477):\penalty0 359--378, 2007.

\bibitem[Gudibande et~al.(2023)Gudibande, Wallace, Snell, Geng, Liu, Abbeel, Levine, and Song]{gudibande2023}
Arnav Gudibande, Eric Wallace, Charlie Snell, Xinyang Geng, Hao Liu, Pieter Abbeel, Sergey Levine, and Dawn Song.
\newblock The false promise of imitating proprietary {LLMs}, 2023.
\newblock arXiv preprint arXiv:2305.15717.

\bibitem[Guo et~al.(2017)Guo, Pleiss, Sun, and Weinberger]{guo2017}
Chuan Guo, Geoff Pleiss, Yu~Sun, and Kilian~Q. Weinberger.
\newblock On calibration of modern neural networks.
\newblock In \emph{Proceedings of the 34th International Conference on Machine Learning}, 2017.

\bibitem[Jitkrittum et~al.(2023)Jitkrittum, Gupta, Menon, Narasimhan, Rawat, and Kumar]{jitkrittum2023}
Wittawat Jitkrittum, Neha Gupta, Aditya~Krishna Menon, Harikrishna Narasimhan, Ankit~Singh Rawat, and Sanjiv Kumar.
\newblock When does confidence-based cascade deferral suffice?
\newblock In \emph{Advances in Neural Information Processing Systems 36}, 2023.

\bibitem[Kumar et~al.(2019)Kumar, Liang, and Ma]{kumar2019}
Ananya Kumar, Percy Liang, and Tengyu Ma.
\newblock Verified uncertainty calibration.
\newblock In \emph{Advances in Neural Information Processing Systems 32}, 2019.

\bibitem[Kwok et~al.(2026)Kwok, Li, Atreya, Liu, Jiang, Finn, Pavone, Stoica, and Mirhoseini]{kwok2026}
Jacky Kwok, Shulu Li, Pranav Atreya, Yuejiang Liu, Yixing Jiang, Chelsea Finn, Marco Pavone, Ion Stoica, and Azalia Mirhoseini.
\newblock {LLM}-as-a-{Verifier}: A general-purpose verification framework, 2026.
\newblock arXiv:2607.05391v2.

\bibitem[Li et~al.(2026)Li, Miao, Krishnan, and Padman]{li2026}
Yubo Li, Yidi Miao, Ramayya Krishnan, and Rema Padman.
\newblock {JEV}-as-a-{Judge}: Accept when confident, escalate when unsure, 2026.
\newblock arXiv:2609.26550v1, 22 September 2026.

\bibitem[Nixon et~al.(2019)Nixon, Dusenberry, Jerfel, Nguyen, Liu, Zhang, and Tran]{nixon2019}
Jeremy Nixon, Mike Dusenberry, Ghassen Jerfel, Timothy Nguyen, Jeremiah Liu, Linchuan Zhang, and Dustin Tran.
\newblock Measuring calibration in deep learning.
\newblock In \emph{CVPR Workshops}, 2019.

\bibitem[Nosek et~al.(2018)Nosek, Ebersole, DeHaven, and Mellor]{nosek2018}
Brian~A. Nosek, Charles~R. Ebersole, Alexander~C. DeHaven, and David~T. Mellor.
\newblock The preregistration revolution.
\newblock \emph{Proceedings of the National Academy of Sciences}, 115\penalty0 (11):\penalty0 2600--2606, 2018.

\bibitem[Pezeshkpour and Hruschka(2024)]{pezeshkpour2024}
Pouya Pezeshkpour and Estevam Hruschka.
\newblock Large language models sensitivity to the order of options in multiple-choice questions.
\newblock In \emph{Findings of the Association for Computational Linguistics: NAACL 2024}, 2024.

\bibitem[Roelofs et~al.(2022)Roelofs, Cain, Shlens, and Mozer]{roelofs2022}
Rebecca Roelofs, Nicholas Cain, Jonathon Shlens, and Michael~C. Mozer.
\newblock Mitigating bias in calibration error estimation.
\newblock In \emph{Proceedings of the 25th International Conference on Artificial Intelligence and Statistics (AISTATS)}, 2022.

\bibitem[Wang et~al.(2024)Wang, Li, Chen, Cai, Zhu, Lin, Cao, Kong, Liu, Liu, and Sui]{wang2024fair}
Peiyi Wang, Lei Li, Liang Chen, Zefan Cai, Dawei Zhu, Binghuai Lin, Yunbo Cao, Lingpeng Kong, Qi~Liu, Tianyu Liu, and Zhifang Sui.
\newblock Large language models are not fair evaluators.
\newblock In \emph{Proceedings of the 62nd Annual Meeting of the Association for Computational Linguistics (Volume 1: Long Papers)}, 2024.

\bibitem[Warner et~al.(2024)Warner, Chaffin, Clavi{\'e}, Weller, Hallstr{\"o}m, Taghadouini, Gallagher, Biswas, Ladhak, Aarsen, Cooper, Adams, Howard, and Poli]{warner2024}
Benjamin Warner, Antoine Chaffin, Benjamin Clavi{\'e}, Orion Weller, Oskar Hallstr{\"o}m, Said Taghadouini, Alexis Gallagher, Raja Biswas, Faisal Ladhak, Tom Aarsen, Nathan Cooper, Griffin Adams, Jeremy Howard, and Iacopo Poli.
\newblock Smarter, better, faster, longer: A modern bidirectional encoder for fast, memory efficient, and long context finetuning and inference, 2024.
\newblock arXiv:2412.13663 (ModernBERT).

\bibitem[Zheng et~al.(2024)Zheng, Zhou, Meng, Zhou, and Huang]{zheng2024mcq}
Chujie Zheng, Hao Zhou, Fandong Meng, Jie Zhou, and Minlie Huang.
\newblock Large language models are not robust multiple choice selectors.
\newblock In \emph{The Twelfth International Conference on Learning Representations}, 2024.

\bibitem[Zheng et~al.(2023)Zheng, Chiang, Sheng, Zhuang, Wu, Zhuang, Lin, Li, Li, Xing, Zhang, Gonzalez, and Stoica]{zheng2023judging}
Lianmin Zheng, Wei-Lin Chiang, Ying Sheng, Siyuan Zhuang, Zhanghao Wu, Yonghao Zhuang, Zi~Lin, Zhuohan Li, Dacheng Li, Eric~P. Xing, Hao Zhang, Joseph~E. Gonzalez, and Ion Stoica.
\newblock Judging {LLM}-as-a-{Judge} with {MT}-{Bench} and {Chatbot} {Arena}.
\newblock In \emph{Advances in Neural Information Processing Systems 36, Datasets and Benchmarks Track}, 2023.

\end{thebibliography}

\appendix

\section{Preregistration document}\label{app:prereg}
The frozen preregistration (frozen 2026-09-26, 15:50~UTC)
specifies: the frozen artifact pins; the Track~A \textsc{calib}/\textsc{eval} case-ID
lists (seed 20260926) and Track~B 960/240 split; the confidence statistic; the
$\tau$ grid, error target ($0.10$), and selection rule; the six calibration methods
and the NLL selection rule; the exact metric definitions (frozen in code with unit
tests); the bootstrap and BCa/FDR specification; hypothesis operationalizations for
H1--H4; the 24-test confirmatory family; the amended local-only comparator set and
oracle-escalation reporting convention; the descoping of the external set; and the
deviations protocol. \emph{[The preregistration file is reproduced verbatim as an
ancillary file in the submission tarball; it is summarized rather than inlined here
to keep the main text within page limits.]}

\section{Ancillary files and the per-decision prediction artifact}\label{app:artifact}
{\raggedright\hbadness=10000
\texttt{released\_full\_test.jsonl} (in \texttt{anc/predictions/}) contains
one JSON object per decision, 2{,}000 rows, with fields: \texttt{case\_id},
\texttt{workflow}, \texttt{question\_id}, \texttt{primitive}, \texttt{n\_options},
\texttt{gold\_argmax}, \texttt{gold\_distribution}, \texttt{pred\_argmax},
\texttt{pred\_distribution}, \texttt{max\_prob}, \texttt{correct},
\texttt{latency\_ms}, and \texttt{provenance} (constant \texttt{"measured"}). The
completed study adds, with the identical schema: \texttt{released\_calib\_half.jsonl}
and \texttt{released\_eval\_half.jsonl} (Track~A halves, raw-logit path),
\texttt{replication\_test.jsonl} (Track~B's single test pass),
\texttt{specialist\_base\_test.jsonl} (ModernBERT-base specialist's single test pass),
\texttt{base\_full\_test.jsonl} and \texttt{multilingual\_full\_test.jsonl} (zero-shot
comparators), and the exploratory probe reruns
\texttt{e1\_fp32\_cpu\_probe.jsonl}, \texttt{e1\_version\_probe\_0\_3\_20.jsonl},
and \texttt{e1\_version\_probe\_0\_3\_3.jsonl} (deviations D10 and D11).
Per-decision JSONs of the frozen-pipeline \textsc{eval} and Track~B
gating decisions (\texttt{e5\_eval\_decisions.json},
\texttt{track\_b\_test\_decisions.json}, \texttt{e4\_calib\_decisions.json}, with
\texttt{specialist\_base\_test\_decisions.json} and
\texttt{track\_b\_validation\_decisions.json}) are in \texttt{anc/metrics/}.\par}

{\raggedright\hbadness=10000
The full ancillary set is: \texttt{anc/preregistration.md} (the frozen plan,
verbatim); \texttt{anc/gate.json}; \texttt{anc/pins.json}; the two split files
(\texttt{track\_a\_split.json}, \texttt{track\_b\_split.json}); the three frozen
calibration pipelines (\texttt{frozen\_pipeline.json},
\texttt{track\_b\_frozen\_pipeline.json},
\texttt{specialist\_base\_frozen\_pipeline.json}); the frozen metric suite with its
unit tests (\texttt{metrics.py}, \texttt{test\_metrics.py});
\texttt{anc/manifests/} (every run manifest, plus the environment pip freeze);
\texttt{anc/predictions/} (the JSONL files above); \texttt{anc/metrics/} (every
canonical metric JSON); \texttt{anc/e10/} (the exploratory probe's corpus,
per-decision predictions, validation record, and results); and
\texttt{anc/README.txt}, which lists the SHA-256 digest of every ancillary file.
The run manifests are released with the author's private source identifiers
removed; no measured field is altered.\par}

\section{Deviations ledger}\label{app:deviations}
Per the preregistration's deviations protocol, all post-freeze deviations are
recorded here. Recorded to date:
\begin{enumerate}
  \item \textbf{2026-09-26, ${\sim}$15:51~UTC (minutes after the freeze):} a
  100-case smoke re-run of the exempt E1 pipeline overwrote the \emph{on-disk}
  copies of the E1 metrics and prediction files. The canonical E1 artifacts are the
  full-400-case versions released in the ancillary files; no analysis uses the
  smoke run. It fits nothing, but it re-read 100 test cases after test-set closure
  and is logged for the exposure ledger. The freeze time is the author's own
  statement, bracketed as stated in \S\ref{sec:prereg}.
  \item \textbf{D1: $\mathrm{ECE}_{\mathrm{sweep}}$ added after the E4
  {\normalfont\textsc{calib}} run started.} The frozen metric suite named the monotonic
  sweep ECE \citep{roelofs2022} but the implementation (with unit tests) landed after
  the E4 \textsc{calib} extraction had begun. It was computed from the stored records;
  no frozen decision (winner, $\tau^{*}$) depends on it.
  \item \textbf{D2: paraphrase test substituted.} The preregistration named no
  runnable paraphraser and none was available under the study's local-only
  constraint; two deterministic surface transforms (lowercase/no-punctuation, neutral
  prefix) ran instead and are labeled as substitutes wherever they appear
  (\S\ref{sec:e7}). No semantic-paraphrase claim is made.
  \item \textbf{D3: canonical E1 artifact restoration.} After the smoke-run
  overwrite (entry 1), the on-disk E1 files were restored from the canonical E1
  files before any E2--E8 analysis and both Track~A halves re-verified
  against the restored file (2{,}000/2{,}000 decisions). {\raggedright\hbadness=10000
  The 2{,}000/2{,}000 cross-check is recomputable from \texttt{anc/predictions/},
  checking \texttt{released\_calib\_half.jsonl} and
  \texttt{released\_eval\_half.jsonl} against \texttt{released\_full\_test.jsonl}.
  The stored \texttt{raw\_logit\_verification} blocks record the earlier 225- and
  275-row checks against the smoke file.\par}
  \item \textbf{D4: step-8 latency re-measurement contended.} The E6 latency run
  executed while Track~B training occupied the same accelerator (p50 $204$~ms vs.\
  $95.9$~ms uncontended); it is retained in the artifact but superseded in the paper
  by the canonical, uncontended E1 measurement, labeled as such (\S\ref{sec:e5}).
  \item \textbf{D5: post-registration exploratory experiment added (E10).} The
  out-of-distribution claim-verification probe of \S\ref{sec:e10} was designed and run
  after the freeze. It touches no benchmark split, enters no confirmatory test, and is
  labeled exploratory on every artifact; it is logged here because it is a
  post-registration addition to the study's scope.
  \item \textbf{D6: correction of a frozen misstatement (Brier target convention).}
  The frozen plan (\S8) attributed the card's $0.062$ Brier to the one-hot form;
  verification against the code of the fine-tuning notebook linked from the model
  card shows it is the vs-teacher form
  (one-hot measures ${\approx}0.37$, vs-teacher $0.0658$). The paper reports the
  correct convention (Table~\ref{tab:t1}); no measured value, selection, or test
  changes; the frozen document is not edited, and this entry is the correction of
  record.
  \item \textbf{D7: card-ECE bin-convention reconciliation.} The preregistration
  froze the card-comparability ECE at 10 bins; manuscript drafts (and one artifact
  key, \texttt{ece\_card\_M15}) described it as 15 bins. The paper now uses the
  frozen 10-bin definition for ``card convention'' and labels equal-width $M{=}15$ as
  the frozen confirmatory endpoint. Numerically invisible on E1; uniform
  under-confidence collapses every binned variant to $0.21423$
  (\S\ref{sec:underconf}); no confirmatory test is affected (the family was always
  defined on equal-width $M{=}15$).
  \item \textbf{D8: p-value convention correction.} The preregistration (\S9) froze
  the raw tail-fraction formula $p = 2\min(P(\Delta^{*} \le 0), P(\Delta^{*} \ge 0))$
  (one-sided where specified), which encodes zero exceedances in $10{,}000$ resamples
  as $p = 0$. At round-1 review this was replaced by the finite-sample add-one
  convention $(b{+}1)/(B{+}1)$, doubled for two-sided tests (\S\ref{sec:stats}).
  \texttt{fdr\_family\_v2.json} stores the corrected values with the raw tail
  fractions preserved per row (\texttt{p\_raw}); the original
  \texttt{fdr\_family.json} is retained unchanged; BH recomputed from the corrected
  p-values changes no decision (20 of 22 executed tests reject; A6 and C6 still fail).
  \item \textbf{D9: post-hoc decomposition of the $\tau^{*}$ miss (2026-09-27).}
  An exploratory analysis added at external review, released as
  \texttt{e11\_tau\_decomposition.json}, recomputes
  accepted-set error at matched coverage under the identity and isotonic rankings,
  sweeps the frozen $\tau$ grid under the single-temperature calibrator, and
  evaluates a counterfactual upper-confidence-bound selection rule. Inputs are stored
  records and grid rows only, all in the ancillary files; no model
  forwards ran, and Track~B's validation side uses stored grid-row aggregates (its
  per-decision records were produced afterwards; D10). No frozen number, verdict,
  threshold, or table changes; the analysis is labeled exploratory and post hoc
  where it appears (\S\ref{sec:e5}).
  \item \textbf{D10: fp32/CPU probe and Track~B validation records (2026-09-27).}
  Two additions at external review, neither changing a frozen number. (i)~An
  exploratory fp32/CPU rerun of the 2{,}000 E1 decisions
  (\texttt{e1\_fp32\_cpu\_probe.json} plus a per-decision file; canonical E1
  artifacts untouched) reproduces the canonical run, F16 weights under bf16
  autocast on MPS (soft accuracy $0.5087$
  vs.\ $0.5086$; max probability difference $0.0079$; 4 argmax flips), ruling out
  precision and backend (\S\ref{sec:e1}). A planned second leg on hosted T4
  hardware could not be run from this environment. (ii)~Track~B validation
  per-decision records (\texttt{track\_b\_validation\_decisions.json}), from a
  forward pass on the 240 validation cases only with the frozen calibration
  parameters reapplied; a sanity gate confirmed the winner's records reproduce the
  stored $\tau^{*}$ grid row exactly ($n = 704$, accepted error $0.0866$), and the
  test set was not touched. These records postdate the D9 analysis.
  \item \textbf{D11: \texttt{laya} version bisection of the soft-accuracy
  discrepancy (2026-09-27).} Exploratory reruns of the 2{,}000 E1 decisions in one
  fresh virtual environment per \texttt{laya} release, on the canonical backend
  (MPS) with torch 2.14.0 and transformers 5.17.0 pinned, so only the harness
  version varies (artifacts \texttt{e1\_version\_probe\_*.json} with per-decision
  files and one manifest per run; canonical artifacts untouched; fits nothing).
  Release 0.3.20 reproduces the canonical run with maximum probability difference
  $0.0$ at the stored rounding; release 0.3.3 (single-state \texttt{predict} API,
  run through a batch-of-one adapter after the first attempt failed on the missing
  \texttt{predict\_batch} method, manifested) measures soft accuracy $0.5087$ with
  4 argmax flips (it differs from the canonical run on the same 1{,}923 decisions
  as the fp32/CPU probe, with the same four flips and the same maximum probability
  difference of $0.0079$; 1{,}910 of the 1{,}923 difference vectors are
  digit-identical at the files' stored precision, consistent with 0.3.3 running
  without autocast). The endpoints agree, so per the pre-stated rule no intermediate
  release was run; the released 0.3.3--0.3.20 harness range is ruled out as the
  cause (\S\ref{sec:e1}).
\end{enumerate}
One provenance note: \texttt{track\_b\_split.json} was written by the training
driver about two minutes after the training manifest opened; the split is
deterministic from seed 20260926 and its contents match the preregistration's
description.

The third-party out-of-distribution report the preregistration anticipated citing
is omitted from this version; \S\ref{sec:e10} stands on the exploratory probe
alone.

\section{Execution timeline}\label{sec:timeline}
The study was executed on one machine (Apple M5~Max), in one automated session on
2026-09-26 and exploratory reruns on 2026-09-27,
with LLM agents carrying out the frozen protocol under the author's direction.
Every timestamp below is UTC, on 2026-09-26 unless dated otherwise, and comes from
a run manifest released in the ancillary files. Events without a manifest of their
own (the freeze, the review rounds) are listed in order without a
seconds-precision timestamp.
\begin{itemize}\setlength{\itemsep}{1pt}
  \item \textbf{09:25:49} environment pinned (\texttt{pins.json}: model, dataset,
  and base-model revisions with checksums; package versions).
  \item \textbf{12:08:55} canonical E1 reproduction run (400 cases, 2{,}000
  decisions; ${\sim}6.4$ minutes wall clock).
  \item \textbf{15:50} preregistration freeze (the frozen document is released
  verbatim as an ancillary file).
  \item \textbf{15:51:06} accidental partial E1 smoke re-run (exposure-ledger entry
  1, Appendix~\ref{app:deviations}).
  \item \textbf{15:58:58} Track~B training start (${\sim}50.8$ minutes); E2--E8
  extraction and fitting ran concurrently on the same machine, which is what
  invalidated the step-8 latency re-measurement (deviation D4).
  \item \textbf{16:06:19} E4 fitting on \textsc{calib}; \textbf{16:13:45} trivial
  comparators; \textbf{16:20:05} the single frozen \textsc{eval} pass (E5/E6,
  ${\sim}8.0$ minutes).
  \item \textbf{16:29:13--16:35} E10 probe, zero-shot comparator passes, and E7/E8
  robustness runs.
  \item \textbf{16:50:59} ModernBERT-base specialist training (${\sim}23.4$
  minutes); \textbf{16:54:20} Track~B single test pass; \textbf{17:14:53} specialist
  single test pass.
  \item \textbf{17:36:41} acceptance gate first evaluated (\texttt{gate.json});
  results locked the same evening.
  \item \textbf{2026-09-26, evening} two rounds of internal review fixes applied.
  \item \textbf{2026-09-27, morning} external-review revision and author-metadata
  fixes applied.
  \item \textbf{2026-09-27, 08:53:09} fp32/CPU rerun of all 400 test cases
  (deviation D10); \textbf{09:04:18} Track~B validation-records pass (240
  validation cases; D10).
  \item \textbf{2026-09-27, 12:39:38} and \textbf{12:46:09} version-bisection
  reruns of all 400 test cases (\texttt{laya} 0.3.20 and 0.3.3; deviation D11; a
  failed 0.3.3 attempt at \textbf{12:44:27} is manifested).
  \item \textbf{2026-09-27} final internal review fixes applied and this version
  prepared.
\end{itemize}
The two rounds of internal review were adversarial reviews produced by two LLM
systems; their full reports are retained
privately by the author. The present revision responds to an
external review received after the first preprint version and was prepared on
2026-09-27; it changes framing, attribution, and prose, changes no frozen number,
and adds the exploratory analyses recorded as deviations D9, D10, and D11.

\section{Confirmatory test family}\label{app:family}
The frozen family (Benjamini--Hochberg FDR at $q = 0.05$; 24 tests): A2--A6
and B2--B6: Track~A \textsc{eval} paired differences in NLL and in equal-width
$M{=}15$ ECE for each calibration method versus identity (two-sided). C2--C6 and
D2--D6: the same on Track~B's test run. E1t/E2t: oracle-scored cascade accuracy
versus oracle-scored random escalation at matched rate on Track~A \textsc{eval} and
Track~B respectively (one-sided). F1t/F2t: external-versus-in-distribution
differences in accuracy and ECE (two-sided; \textbf{not executed in this version},
external set descoped, \S\ref{sec:e9}; conservatively kept in the BH denominator,
$m = 24$). All other analyses are exploratory and are labeled as such wherever they
appear. Table~\ref{tab:family} gives every executed outcome (all \meas{}); 20 of 22
executed tests reject. P-values carry the add-one correction of \S\ref{sec:stats};
\texttt{fdr\_family\_v2.json} stores the corrected values with raw tail fractions
preserved per row, and recomputing BH from them changes no decision (the original
\texttt{fdr\_family.json} is retained unchanged).

\begin{table}[h]
\centering\footnotesize
\setlength{\tabcolsep}{4pt}
\begin{tabular}{llrrl}
\toprule
ID & Contrast (vs.\ identity unless noted) & $\Delta$ & $p$ & BH outcome \\
\midrule
A2/B2 & \textsc{eval} NLL / ECE, single temp.        & $-0.166$ / $-0.166$ & ${\le}2{\times}10^{-4}$ / ${\le}2{\times}10^{-4}$ & reject / reject \\
A3/B3 & \textsc{eval} NLL / ECE, per-primitive       & $-0.167$ / $-0.164$ & ${\le}2{\times}10^{-4}$ / ${\le}2{\times}10^{-4}$ & reject / reject \\
A4/B4 & \textsc{eval} NLL / ECE, per-option-count    & $-0.167$ / $-0.164$ & ${\le}2{\times}10^{-4}$ / ${\le}2{\times}10^{-4}$ & reject / reject \\
A5/B5 & \textsc{eval} NLL / ECE, vector              & $-0.168$ / $-0.172$ & ${\le}2{\times}10^{-4}$ / ${\le}2{\times}10^{-4}$ & reject / reject \\
A6/B6 & \textsc{eval} NLL / ECE, isotonic (winner)   & $+0.071$ / $-0.153$ & $0.446$ / ${\le}2{\times}10^{-4}$ & \textbf{fail} / reject \\
C2/D2 & Track~B NLL / ECE, single temp.              & $-0.097$ / $-0.129$ & ${\le}2{\times}10^{-4}$ / ${\le}2{\times}10^{-4}$ & reject / reject \\
C3/D3 & Track~B NLL / ECE, per-primitive             & $-0.096$ / $-0.127$ & ${\le}2{\times}10^{-4}$ / ${\le}2{\times}10^{-4}$ & reject / reject \\
C4/D4 & Track~B NLL / ECE, per-option-count          & $-0.096$ / $-0.127$ & ${\le}2{\times}10^{-4}$ / ${\le}2{\times}10^{-4}$ & reject / reject \\
C5/D5 & Track~B NLL / ECE, vector                    & $-0.086$ / $-0.133$ & ${\le}2{\times}10^{-4}$ / ${\le}2{\times}10^{-4}$ & reject / reject \\
C6/D6 & Track~B NLL / ECE, isotonic (winner)         & $-0.046$ / $-0.127$ & $0.108$ / ${\le}2{\times}10^{-4}$ & \textbf{fail} / reject \\
E1t   & \textsc{eval} cascade vs.\ random escalation & $+0.0646$ & ${\le}10^{-4}$ & reject \\
E2t   & Track~B cascade vs.\ random escalation       & $+0.0757$ & ${\le}10^{-4}$ & reject \\
F1t/F2t & external set (descoped)                    & n/a & not executed & not executed \\
\bottomrule
\end{tabular}
\caption{\textbf{Confirmatory family outcomes} (\meas{}; BH-FDR $q{=}0.05$, $m{=}24$).
Paired case-cluster bootstrap deltas; $p \le 2{\times}10^{-4}$ (two-sided) and
$p \le 10^{-4}$ (one-sided, E1t/E2t) are add-one-corrected bounds,
$(b{+}1)/(B{+}1)$ with $b{=}0$ exceedances in $B{=}10{,}000$ resamples (the
resolution limit). Hypothesis verdicts as operationalized: \textbf{H1} supported
except soft accuracy (\S\ref{sec:e1}); \textbf{H2 not supported on either track} (the
frozen selection rule's winner fails its NLL contrast; A6, C6); \textbf{H3 partially
supported on both tracks} (b: better-than-random gating holds, E1t/E2t; a: the $0.10$
error target fails out-of-sample); \textbf{H4 not tested} (descoped).}
\label{tab:family}
\end{table}

\begin{figure}[h]
\centering
\includegraphics[width=0.5\linewidth]{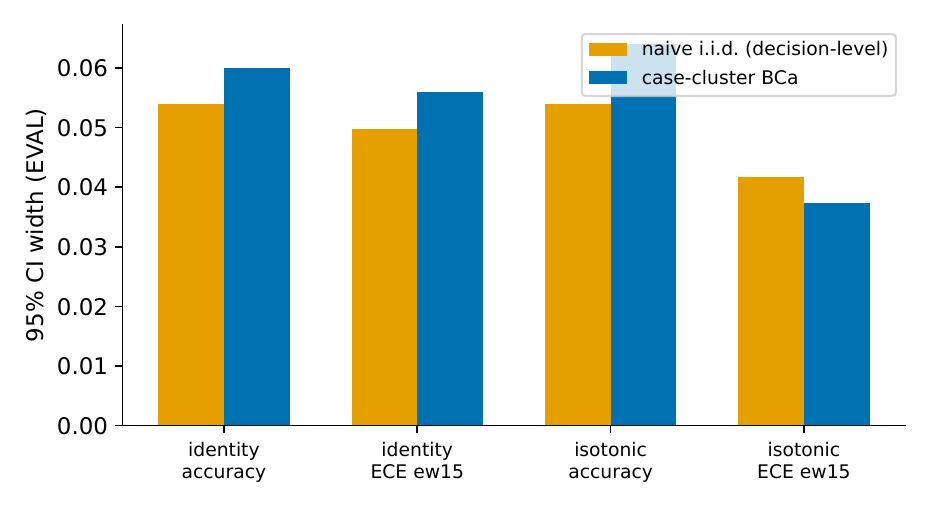}
\caption{\textbf{F8} (\meas{}, appendix). Naive i.i.d.\ decision-level vs.\
case-cluster bootstrap interval widths on \textsc{eval} headline metrics. On winner
accuracy the clustered interval is $1.185\times$ wider; decision-level resampling
understates uncertainty on this five-decisions-per-case benchmark. (Clustered
intervals are BCa, naive are percentile, so the one reversal (isotonic ECE) may
reflect the interval method rather than clustering.)}
\label{fig:f8}
\end{figure}

\end{document}